\documentclass{article}

\usepackage[OT1]{fontenc}
\usepackage{iclr2027_conference,times}

\usepackage{amsmath,amsfonts,bm}

\def\eqref#1{equation~\ref{#1}}

\def\1{\bm{1}}

\DeclareMathAlphabet{\mathsfit}{\encodingdefault}{\sfdefault}{m}{sl}
\SetMathAlphabet{\mathsfit}{bold}{\encodingdefault}{\sfdefault}{bx}{n}

\usepackage{amsmath,amssymb,mathtools}
\usepackage{booktabs,multirow,array,graphicx}
\usepackage{xcolor}
\usepackage{colortbl}
\usepackage{wrapfig}
\usepackage{microtype}
\usepackage{hyperref}
\usepackage{enumitem}
\usepackage{url}
\usepackage{needspace}
\usepackage{capt-of}
\usepackage{tcolorbox}
\tcbuselibrary{skins,breakable}

\hypersetup{hidelinks}

\iclrfinalcopy

\usepackage{fancyhdr}
\usepackage{fontawesome5} 

\newcommand{\paperdate}{September 28, 2026}

\fancypagestyle{paperfirstpage}{%
  \fancyhf{}%

  \fancyhead[L]{%
    \includegraphics[height=20pt]{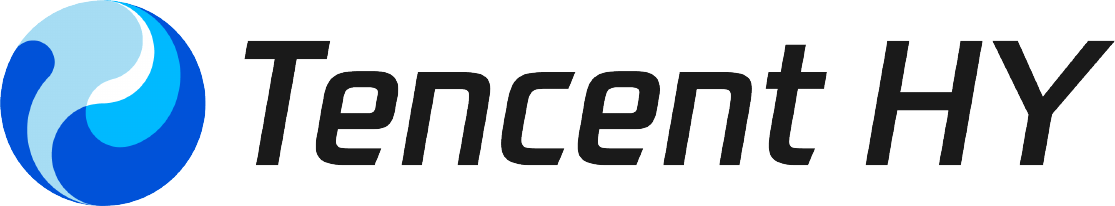}%
  }%

  \fancyhead[R]{%
    \normalfont\small
    \textcolor[HTML]{4D4D4D}{\paperdate}%
  }%

  \fancyfoot[C]{\thepage}%

  \renewcommand{\headrulewidth}{0.5pt}%
  \renewcommand{\footrulewidth}{0pt}%
}

\hypersetup{
  pdfauthor={Xu Wang, Yifan Yang, Tinghao Yu, Difan Zou},
  pdftitle={Beyond Token Scale: Chunk-Level Sparse Autoencoders
    for Reliable Semantic Feature Discovery}
}

\newcommand{\rfve}{\mathrm{RFVE}}
\newcommand{\fve}{\mathrm{FVE}}

\newcommand{\safegraphic}[2]{%
  \IfFileExists{#1}{%
    \includegraphics[width=#2]{#1}%
  }{%
    \fbox{%
      \parbox[c][0.16\textheight][c]{0.94\linewidth}{%
        \centering\small
        Figure unavailable%
      }%
    }%
  }%
}

\definecolor{chunkshade}{HTML}{F1F3F5}

\definecolor{paperNavy}{HTML}{0C445E}
\definecolor{paperBurgundy}{HTML}{893A37}

\definecolor{promptblue}{HTML}{F3F8FB}
\definecolor{promptline}{HTML}{99BAD1}
\definecolor{prompttitle}{HTML}{0C445E}
\definecolor{prompttext}{HTML}{252932}

\newtcolorbox{promptbox}[1][]{%
  enhanced jigsaw,
  breakable,
  width=\linewidth,
  colback=promptblue,
  colframe=promptline,
  coltext=prompttext,
  colbacktitle=promptblue,
  coltitle=prompttitle,
  fonttitle=\normalfont\small\bfseries,
  fontupper=\normalfont\small,
  boxrule=0.5pt,
  titlerule=0.4pt,
  arc=1.5pt,
  left=7pt,
  right=7pt,
  top=6pt,
  bottom=6pt,
  before skip=8pt,
  after skip=8pt,
  before upper={%
    \setlength{\parindent}{0pt}%
    \setlength{\parskip}{3pt}%
  },
  #1
}

\title{Beyond Token Scale: Chunk-Level Sparse \\
Autoencoders for Reliable Semantic \\
Feature Discovery}

\author{%
  \textbf{Xu Wang}\textsuperscript{1,2}\quad
  \textbf{Yifan Yang}\textsuperscript{2}\quad
  \textbf{Tinghao Yu}\textsuperscript{2}\quad
  \textbf{Difan Zou}\textsuperscript{1}\\[6pt]
  \normalfont
  \textsuperscript{1}The University of Hong Kong\quad
  \textsuperscript{2}Hunyuan Team, Tencent\\[4pt]
  \normalfont\small
  \makebox[0pt][l]{%
    \raisebox{-14pt}[0pt][0pt]{%
      \normalfont\footnotesize
      \href{https://github.com/Xu0615/Chunk_Level_SAE}{%
        \faGithub\hspace{0.4em}%
        \nolinkurl{github.com/Xu0615/Chunk_Level_SAE}%
      }%
    }%
  }%
  \mbox{%
    \{%
    \href{mailto:sunny615@connect.hku.hk}{sunny615@connect.hku.hk},
    \href{mailto:dzou@hku.hk}{dzou@hku.hk}%
    \}\quad
    \{%
    \href{mailto:ioanyang@tencent.com}{ioanyang},
    \href{mailto:maxwellyu@tencent.com}{maxwellyu}%
    \}@tencent.com%
  }
}

\begin{document}

\maketitle

\pagestyle{plain}
\thispagestyle{paperfirstpage}


\begin{abstract}
Sparse autoencoders (SAEs) expose features that help us understand and steer language models, but faithful reconstruction does not guarantee informative concepts. Token-level objectives reward lexical and formatting details alongside semantic content, all competing for a limited sparse budget. We introduce a family of chunk-level SAEs that encode mean-pooled activations over chunks, each a contiguous span of tokens: Mean-Chunk reconstructs the observed chunk, Cross-Chunk predicts an independently processed neighbor, and Joint-Chunk combines both targets. These designs separate the effect of a larger observation unit from that of predicting information shared across passages. With matched training data, chunk-level SAEs remain powerful interpretability tools while learning reliable semantic features that capture high-level concepts and respond selectively to relevant content. Their strengths are complementary: Mean-Chunk improves high-level feature discovery, reasoning detection beyond surface cues, and steering; Cross-Chunk leads document retrieval and classification transfer while producing selective, persistent features. Changing what an SAE sees and predicts yields reliable semantic features for more meaningful tasks. We demonstrate their practical value through gains across downstream tasks such as retrieval, reasoning detection, and steering.
\end{abstract}

\begin{figure}[h]
  \centering
  \includegraphics[width=1.0\linewidth]{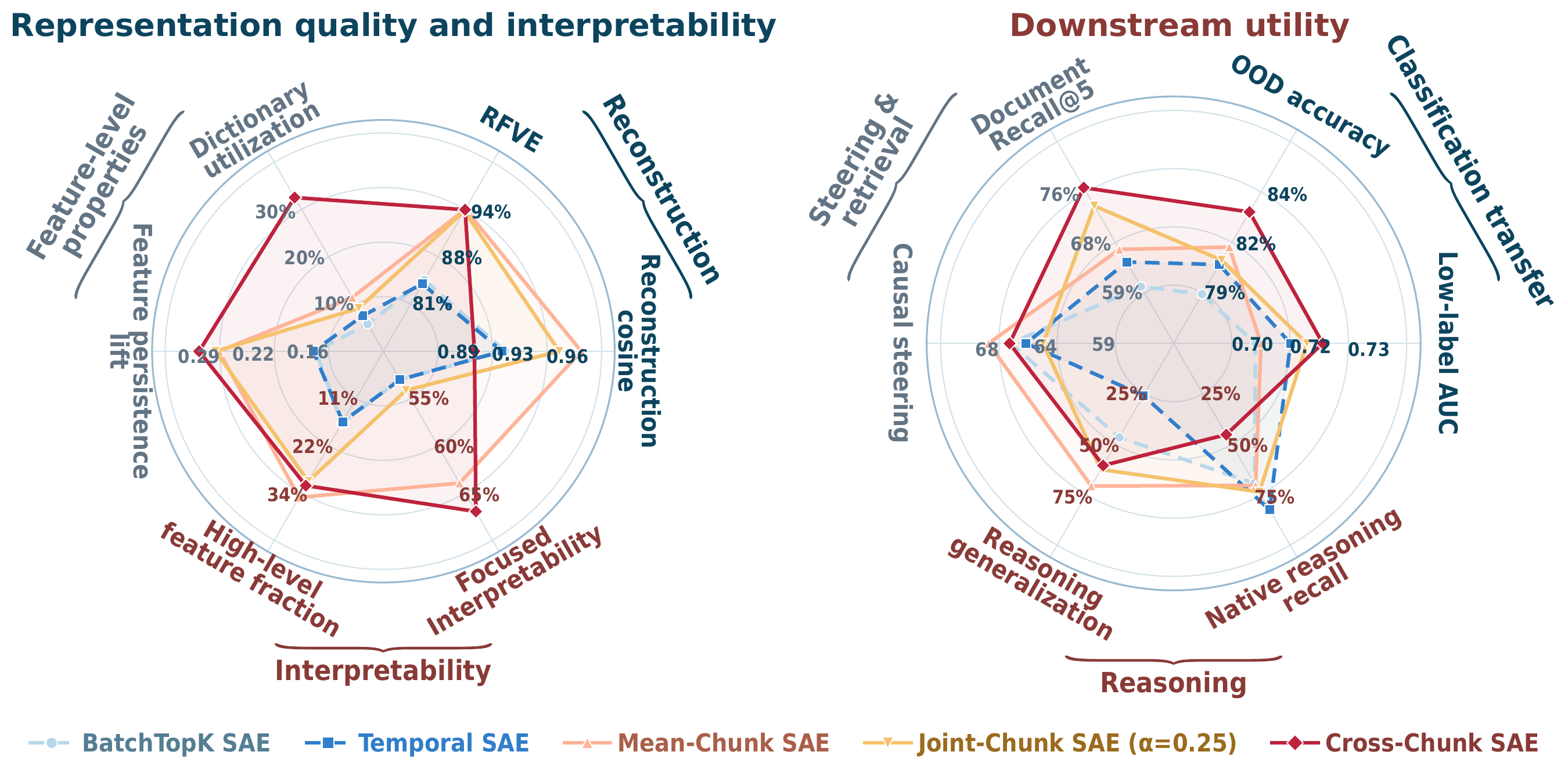}
\caption{
\textbf{SAE representation quality and downstream utility.}
Left: reconstruction fidelity (activation preservation), semantic interpretability (concept clarity and abstraction), feature persistence (consistency across related contexts), and dictionary utilization (breadth of feature use).
Right: retrieval (semantic matching), classification transfer (domain generalization and label efficiency), reasoning detection (sensitivity beyond surface cues), and steering (causal output control).
\textbf{Chunk-level SAEs learn more meaningful features with complementary strengths across downstream tasks.}
}
  \label{fig:summary}
\end{figure}

\section{Introduction}
\label{sec:intro}

Language models are remarkably capable, yet understanding how they arrive at their outputs remains difficult: information is distributed across many neurons, whose individual activities often have no clear meaning. Sparse autoencoders (SAEs) offer a more interpretable view by decomposing activations into sparse combinations of features~\citep{bricken2023monosemanticity, cunningham2023sparse,templeton2024scaling}. These features provide concrete handles for editing causal circuits~\citep{marks2024sparsefeatures}, selectively unlearning knowledge~\citep{farrell2024unlearning,wang-etal-2025-model-unlearning}, and steering instruction-following behavior~\citep{he2025saif}. The appeal of SAEs is therefore practical: revealing human-readable concepts we can inspect and use to understand or steer model behavior.

In addition to the standard SAE, a number of different architectures
have been recently developed, such as Gated~\citep{rajamanoharan2024gated},
JumpReLU~\citep{rajamanoharan2024jumprelu}, TopK~\citep{gao2024scaling},
BatchTopK~\citep{bussmann2024batchtopk}, Matryoshka~\citep{bussmann2025matryoshka},
and Temporal~\citep{bhalla2026temporalsae}, to further improve reconstruction
under a limited sparsity budget through better gating, feature selection,
and dictionary organization. These \emph{token-level SAEs} encode individual
token states and retain token reconstruction as their common target.
However, better reconstruction may not guarantee better
semantic features~\citep{karvonen2025saebench,chanin2025sparsebutwrong}:
architectural improvements remain centered on reconstruction loss, while
semantic feature quality is less directly constrained.
A concrete example has been observed for reasoning feature detection via
SAE~\citep{ma2026sparseautoencodersidentifyreasoning}:
``\texttt{The route is blocked; let's return and try another}''
expresses backtracking without ``\texttt{wait},'' whereas
``\texttt{Please wait by the entrance}'' contains the cue
without any backtracking.
Therefore, a semantically coherent feature should recognize the
revision of a plan across different wordings and reject isolated
surface cues.

Features that track meaning across different wordings need to capture more than
isolated lexical cues, making longer spans a natural unit for sparse coding.
Recent work takes steps in this direction: turn-averaged SAEs pool activations
to discover high-level features in multi-turn dialogue~\citep{der2026turnaveraged},
while step-level SAEs reconstruct reasoning steps conditioned on prior
context~\citep{yang2026steplevel}. These approaches motivate moving beyond
individual tokens, but a broader input alone does not determine what features
will learn. Reconstructing a pooled representation still rewards any words
or formatting preserved in the average, whereas predicting a related passage
may favor information that carries across different wordings. We therefore
study two distinct choices together in ordinary text: how much text should
a feature encode, and what should it predict?

These two questions naturally lead to our method: a family of
\emph{chunk-level SAEs} that varies both the scale of observation and the
target of prediction. A chunk is a contiguous span of tokens, and we train
on pairs of non-overlapping chunks of varying lengths from the same document,
processing each independently. Moreover, we consider three different
chunk-level SAEs, each designed to preserve different information:
Mean-Chunk reconstructs a chunk's own mean activation, Cross-Chunk predicts
the mean activation of its same-document neighbor, and Joint-Chunk combines
both targets through a nested shared dictionary. Using the same sparse-coding
backbone and token stream across multiple chunk lengths lets us separate
the benefits of pooling from those of predicting a neighboring passage. Together, these designs show how the input span and prediction target shape what features learn and how selectively they respond to content across passages.

\textbf{The resulting dictionaries retain strong target-relative fidelity while uncovering reliable semantic features, making SAEs more effective interpretability tools} (see Figure~\ref{fig:summary}). These features capture meaningful concepts across different wordings, with activation patterns that follow relevant content while largely rejecting unrelated cues. The difference becomes particularly clear in the inspected passage traces. On a news passage, token-level SAEs can activate several unrelated features, obscuring which aspects of the text they represent. Cross-Chunk instead produces a cleaner pattern, with activity concentrated in features relevant to the passage and unrelated features remaining largely inactive. This makes it easier to connect a feature's explanation to its observed behavior in context. Chunk-level training thus improves both the semantic content of the dictionary and the clarity of its responses. We summarize the contributions of this paper as follows:

\begin{itemize}[leftmargin=*,nosep]
\item We introduce a family of chunk-level SAEs for semantic feature discovery.
Mean-Chunk reconstructs a span's mean activation, Cross-Chunk predicts an
independently processed neighbor, and Joint-Chunk combines the two through
a nested dictionary. A shared sparse-coding backbone, token stream, and
range of chunk lengths make the roles of pooling and partner prediction explicit.

\item Chunk-level training improves semantic feature quality and cross-passage consistency.
In a random sample of dictionary coordinates, Mean-Chunk \textbf{more than doubles
the high-level feature} fraction to 34.8\% from BatchTopK's 17.2\%,
while retaining strong target-relative fidelity. Across related passages,
Cross-Chunk achieves the strongest feature persistence and over five times
BatchTopK's dictionary utilization, alongside more human-readable explanations
than the token-level baselines.

\item We further demonstrate that the improvements of Chunk-level SAEs can be extended to downstream tasks that restrict reliance on surface cues.
In retrieval with little lexical overlap, Cross-Chunk reaches 77.0\% Recall@5
versus 64.1\% for Temporal, the strongest token-level baseline.
Even after reasoning cues are removed, Mean-Chunk detects backtracking
with 66\% recall versus under 5\% for both token baselines,
\textbf{a gain of over 60 percentage points}.
Cross-Chunk also leads classification transfer, while Mean-Chunk achieves
the highest steering.
\end{itemize}
\begin{figure}[t]
  \centering
  \includegraphics[width=0.9\linewidth]{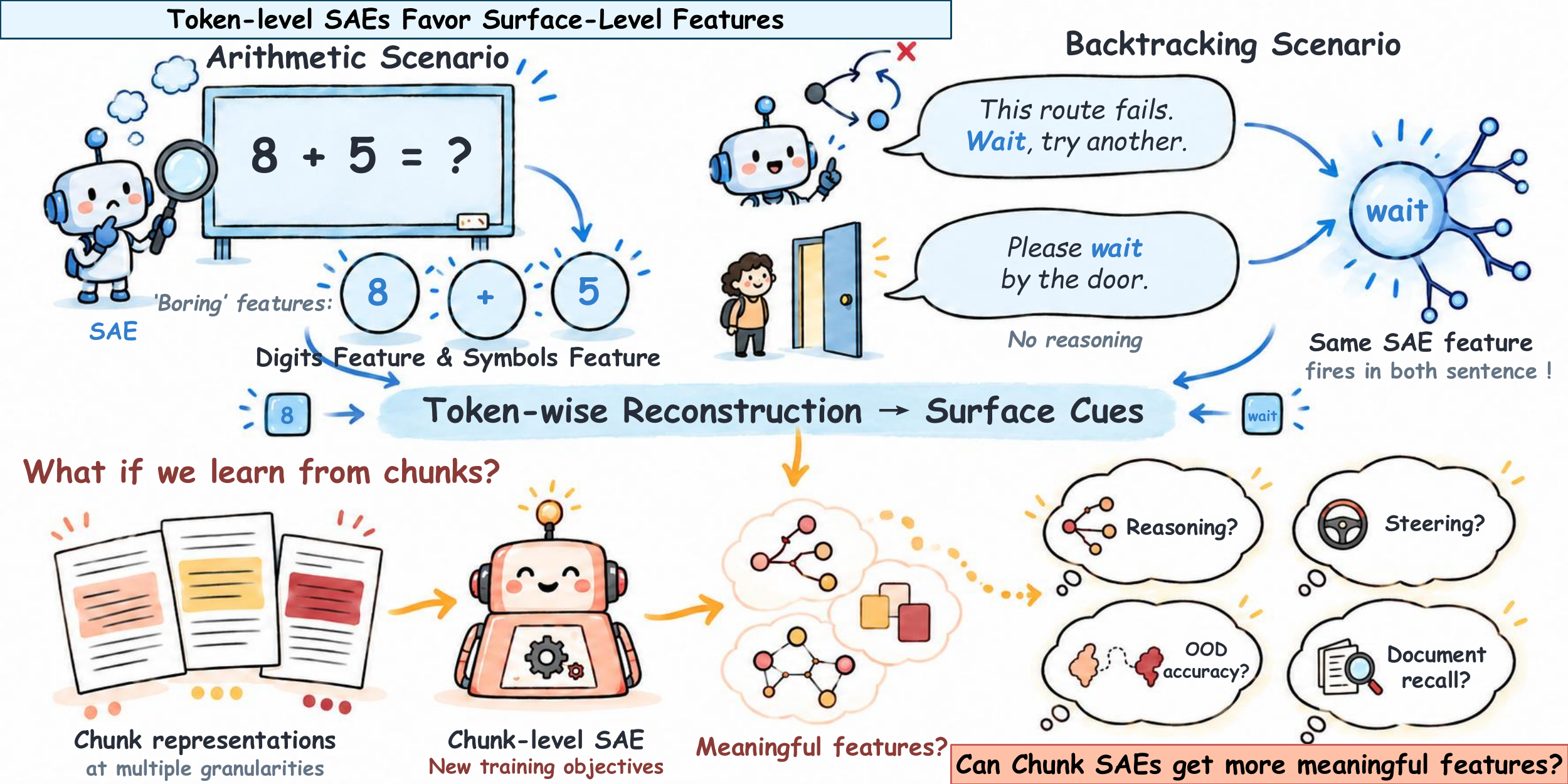}
  \vskip -.1in
  \caption{\textbf{Rethinking token-level sparse autoencoders.}
  Top left: An arithmetic example illustrates we often get 'boring' features.
  Top right: The same feature activates on ``wait'' in both backtracking and non-reasoning contexts.
  Bottom: We investigate whether modifying input granularity and training objectives yields more meaningful SAE features and validate across diverse downstream applications.}
  \label{fig:overview}
\end{figure}

\section{Token-Level SAEs Favor Surface-Level Features}
\label{sec:tokensae}

SAEs use dictionary learning to expose interpretable directions in model
activations~\citep{olshausen1996emergence,elhage2022toy}.
We use BatchTopK and Temporal SAEs as representative token-level baselines:
both reconstruct individual token states under a limited sparse budget,
while Temporal additionally encourages consistency between neighboring codes.

\textbf{A shared sparse-coding backbone.}
For an input $\mathbf{x}\in\mathbb{R}^{d}$, the encoder computes
$\mathbf{a}(\mathbf{x})
=\operatorname{ReLU}(W_{\rm enc}(s\mathbf{x}-\mathbf{b}_{\rm in})
+\mathbf{b}_{\rm enc})$.
The sparse code and decoded prediction are
\begin{equation}
\mathbf{z}(\mathbf{x})
=\operatorname{BatchTopK}_{K}(\mathbf{a}(\mathbf{x})),
\qquad
\widehat{\mathbf{y}}(\mathbf{x})
=W_{\rm dec}\mathbf{z}(\mathbf{x})+\mathbf{b}_{\rm out}.
\label{eq:shared}
\end{equation}
Here, $W_{\rm enc}\in\mathbb{R}^{m\times d}$,
$W_{\rm dec}\in\mathbb{R}^{d\times m}$, and $s$ fixes the activation scale.
BatchTopK retains the largest positive activations under a batch-wide
budget of $nK$ for $n$ inputs~\citep{bussmann2024batchtopk}.
All methods normalize decoder columns and use squared reconstruction error
$\mathcal{L}_{\rm rec}(\mathbf{x},\mathbf{y})
=\|\widehat{\mathbf{y}}(\mathbf{x})-s\mathbf{y}\|_2^2$.
The shared AuxK penalty revives inactive features with weight
$\lambda_{\rm aux}=0.0625$~\citep{gao2024scaling}.

\textbf{Token-level training objectives.}
BatchTopK reconstructs each contextualized token state $\mathbf{h}_t$.
Temporal adds prefix reconstruction and symmetric contrastive alignment
between adjacent-token prefix codes~\citep{bhalla2026temporalsae,oord2018cpc}:
\begin{align}
\mathcal{L}_{\rm BatchTopK}
&=\mathbb{E}_{t}
[\mathcal{L}_{\rm rec}(\mathbf{h}_t,\mathbf{h}_t)]
+\lambda_{\rm aux}\mathcal{L}_{\rm AuxK},
\label{eq:batchtopk-main}\\
\mathcal{L}_{\rm Temporal}
&=\frac{0.8\mathcal{L}_{\rm full}+0.2\mathcal{L}_{H}}{2}
+\mathcal{L}_{\rm InfoNCE}^{\rm sym}
+\lambda_{\rm aux}\mathcal{L}_{\rm AuxK}.
\label{eq:temporal-main}
\end{align}
Here, $\mathbb{E}_{t}$ is computed by averaging the reconstruction loss
over all non-padding token positions in each training minibatch,
with equal weight for each position.
The full-code loss $\mathcal{L}_{\rm full}$ uses the same averaging rule.
The prefix loss $\mathcal{L}_{H}$ instead averages over tokens with a
preceding token in the same input sequence, measuring how well a
designated subset of features reconstructs the current token.
The contrastive term encourages this subset to carry similar information
across neighboring tokens. Temporal therefore changes how features
relate across tokens, but both baselines still learn to reproduce
individual token states. This objective rewards surface details
alongside semantic content, motivating our shift toward observing
and predicting longer spans. See Appendix~\ref{app:baseline-objectives} for details.

\section{Designing Chunk-Level SAEs to Learn Meaningful Features}
\label{sec:chunksae}

Our design assigns one sparse code to a text span rather than to each token.
We form adjacent, non-overlapping chunks $A$ and $B$ from the same document
and process them independently, resetting position indices and blocking
attention between them. For a chunk $A$ containing $L_A$ tokens, its mean
activation is
$\boldsymbol{\mu}_A=L_A^{-1}\sum_{t=1}^{L_A}\mathbf{h}^{A}_t$,
where $\mathbf{h}^{A}_t\in\mathbb{R}^{d}$ is the layer-$\ell$ activation
at position $t$; $\boldsymbol{\mu}_B$ is defined analogously.
We vary both chunk lengths to cover different context scales and encode
each mean using Eq.~\ref{eq:shared}. This gives us a common starting point
for comparing three training targets: reconstructing the observed chunk,
predicting its neighbor, and combining both tasks.

\subsection{Mean-Chunk: Reconstructing the Chunk Mean}

The first step is to give each sparse code a broader view of the text
without changing the reconstruction task. Averaging token activations
lets the code describe a span as a whole, rather than spend a separate
sparse budget at every position. Mean-Chunk therefore encodes the
pooled activation $\boldsymbol{\mu}_A$ and learns to reconstruct
that same mean:
\begin{equation}
\mathcal{L}_{\rm mean}
=\mathbb{E}_{A}
[\mathcal{L}_{\rm rec}(\boldsymbol{\mu}_A,\boldsymbol{\mu}_A)]
+\lambda_{\rm aux}\mathcal{L}_{\rm AuxK}.
\label{eq:mean}
\end{equation}
Here, $\mathbb{E}_{A}$ averages over chunks produced by the shared
training-pair sampler, using the chunk lengths specified in
Appendix~\ref{app:protocol}. In each minibatch, we average the
per-chunk reconstruction losses, so each sampled chunk contributes
one loss regardless of its length.
The reconstruction error $\mathcal{L}_{\rm rec}$ is defined in
Section~\ref{sec:tokensae}, and the shared AuxK penalty revives inactive
features with weight $\lambda_{\rm aux}=0.0625$.
This design isolates the effect of pooling, which preserves both semantic content and surface details.

\subsection{Cross-Chunk: Predicting a Same-Document Neighbor}

To change that priority, Cross-Chunk asks the code to predict a
neighboring passage instead of reconstructing its own input.
Two passages may describe the same event in different words, so
their shared subject can help predict the neighbor even when
passage-specific details cannot. We implement this idea by replacing
the self target in Eq.~\ref{eq:mean} with the neighboring chunk's mean
and averaging both prediction directions:
\begin{equation}
\mathcal{L}_{\rm cross}
=\tfrac{1}{2}\mathbb{E}_{A,B}
\big[
\mathcal{L}_{\rm rec}(\boldsymbol{\mu}_A,\boldsymbol{\mu}_B)
+\mathcal{L}_{\rm rec}(\boldsymbol{\mu}_B,\boldsymbol{\mu}_A)
\big]
+\lambda_{\rm aux}\mathcal{L}_{\rm AuxK}.
\label{eq:cross}
\end{equation}
The expectation averages over neighboring chunk pairs, with the factor
$1/2$ equally weighting prediction from $A$ to $B$ and from $B$ to $A$.
Both directions use the same encoder and decoder, so each chunk serves
as both input and target. Independent processing prevents the encoder
from reading the target passage, encouraging it to capture information
in the input that helps predict its neighbor. Shared topics can therefore
be more useful than passage-specific wording.

\subsection{Joint-Chunk: Balancing Self and Partner Targets}

Joint-Chunk encodes the observed chunk mean $\boldsymbol{\mu}_A$
once, producing $\mathbf{z}_A=\mathbf{z}(\boldsymbol{\mu}_A)$ with
dictionary width $m=65{,}536$ and batch-average sparse budget
$K=128$. Its two predictions are
\begin{equation}
\widehat{\mathbf{y}}_{\rm self}(\boldsymbol{\mu}_A)
=W_{\rm dec}\mathbf{z}_A+\mathbf{b}_{\rm self},
\qquad
\widehat{\mathbf{y}}_{\rm partner}(\boldsymbol{\mu}_A)
=W_{\rm dec}[:,1:h]\mathbf{z}_{A,1:h}
+\mathbf{b}_{\rm partner},
\quad h=32{,}768.
\label{eq:joint-decode}
\end{equation}
The self decoder reads the full sparse code, while the partner decoder
reads only its shared prefix. Prefix coordinates therefore share decoder
columns across the two predictions; the output biases are separate.
The prefix is sliced after a single BatchTopK operation.

Let $b_{\rm self}$ and $b_{\rm partner}$ denote the training-set
mean constant-predictor squared errors for the two targets, as
defined in Eq.~\ref{eq:joint-normalizers}. For a directed pair
$A\rightarrow B$, the reconstruction loss is
\begin{equation}
\ell_{\rm joint}(A,B;\alpha)
=\frac{1}{1+\alpha}
\left[
\frac{
\|\widehat{\mathbf{y}}_{\rm self}(\boldsymbol{\mu}_A)
-s\boldsymbol{\mu}_A\|_2^2
}{b_{\rm self}}
+\alpha
\frac{
\|\widehat{\mathbf{y}}_{\rm partner}(\boldsymbol{\mu}_A)
-s\boldsymbol{\mu}_B\|_2^2
}{b_{\rm partner}}
\right].
\label{eq:joint-task}
\end{equation}
Each prediction error is divided by the constant-predictor normalizer
for its own target. The shared activation scale $s$ applies to both
targets, and $\alpha$ sets the partner term's weight relative to the
self term. The factor $1/(1+\alpha)$ ensures that the two task weights
sum to one.

We combine this loss with the per-pair AuxK residual loss
defined in Eq.~\ref{eq:joint-aux-per-pair}:
\begin{equation}
\mathcal{L}_{\rm joint}(\alpha)
=\mathbb{E}_{(A,B)}
\left[
\ell_{\rm joint}(A,B;\alpha)
+\lambda_{\rm aux}\ell_{\rm AuxK}(A,B;\alpha)
\right],
\qquad
\lambda_{\rm aux}=0.0625.
\label{eq:joint}
\end{equation}
The AuxK term uses the same target normalizers and task weights as
Eq.~\ref{eq:joint-task}, keeping its scale aligned with the
reconstruction terms. The expectation averages both terms over
directed training pairs. We use $\alpha=0.25$ in the main comparisons
and evaluate $\alpha\in\{0.5,1,1.5\}$ in
Appendix~\ref{app:ablations}.

\section{Chunk-Level SAEs Reveal Clearer Semantic Structure}
\label{sec:faith_interp}
\begingroup
\setlength{\parfillskip}{0pt}
\tolerance=1000
\emergencystretch=0pt
\makeatletter
\c@topnumber=1
\c@bottomnumber=1
\c@totalnumber=1
\renewcommand{\bottomfraction}{0.6}
\@floatplacement
\aftergroup\@floatplacement
\makeatother

We test whether chunk-level codes retain information, expose
concepts, and respond selectively across related passages.
These analyses connect reconstruction fidelity to semantic feature quality.

\begingroup
\setlength{\parfillskip}{0pt plus 1fil}
\sloppy
\begin{figure}[t]
  \centering
  \includegraphics[width=.9\linewidth]{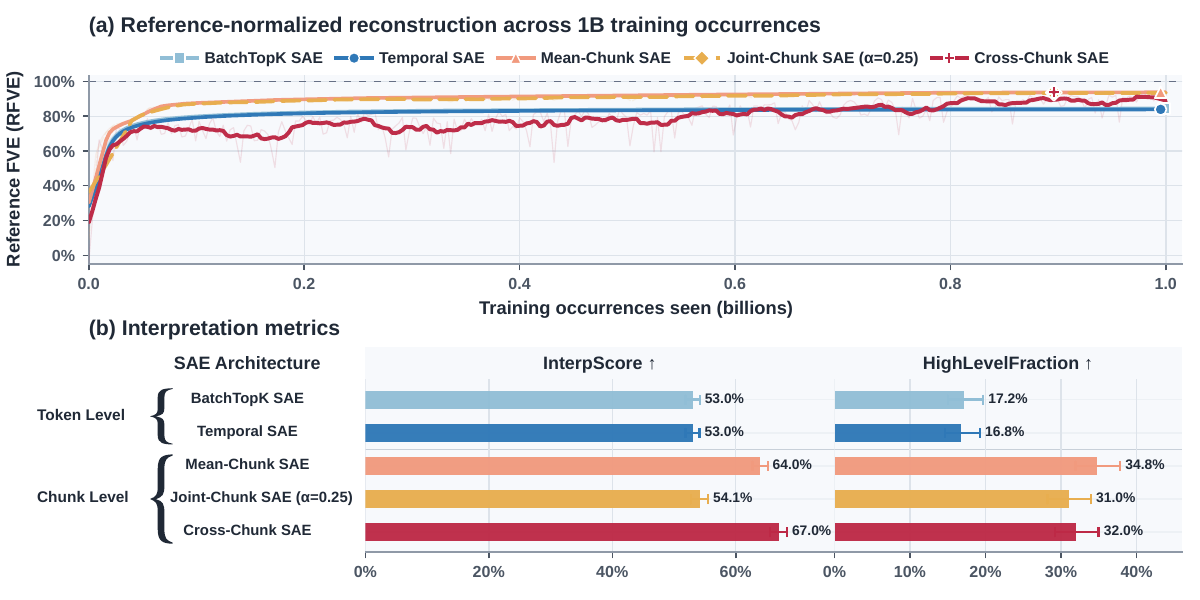}
  \vskip -.1in
\caption{\textbf{Training fidelity and feature interpretation.}
  Top (a): RFVE during training, normalized to a target-appropriate reference.
  Bottom (b): architectures grouped by encoding granularity, focused interpretability (InterpScore), and high-level feature fraction (HighLevelFraction).
 \textbf{Chunk-level SAEs improve semantic feature quality while retaining strong target-relative fidelity.}}
  \label{fig:faith_interp}
\end{figure}
\endgroup

\subsection{Experimental Setup}

All five SAE families use layer-21 activations from
Qwen3.5-9B-Base~\citep{qwen2026qwen35}, with $d=4096$,
$m=65{,}536$, and BatchTopK budget $K=128$. Training uses the same
one billion Pile token occurrences~\citep{gao2020pile,lee2022deduplicating}
and chunk lengths $\{32,64,128,256,512\}$. Paired chunks are
non-overlapping and processed independently with position indices
reset. Joint-Chunk uses $\alpha=0.25$; Appendix~\ref{app:ablations}
examines additional partner weights under the same training protocol.

Token-level SAEs encode and threshold tokens before averaging their
codes; chunk-level SAEs average hidden states before encoding and
thresholding. Reasoning instead uses document-level scoring
(Appendix~\ref{app:prompts}). Training settings, splits, and metric
definitions are collected in Appendices~\ref{app:protocol}
and~\ref{app:metric-definitions}.

\begingroup
\setlength{\parfillskip}{0pt plus 1fil}
\sloppy
\begin{figure}[t]
  \centering
  \includegraphics[width=1.0\linewidth]{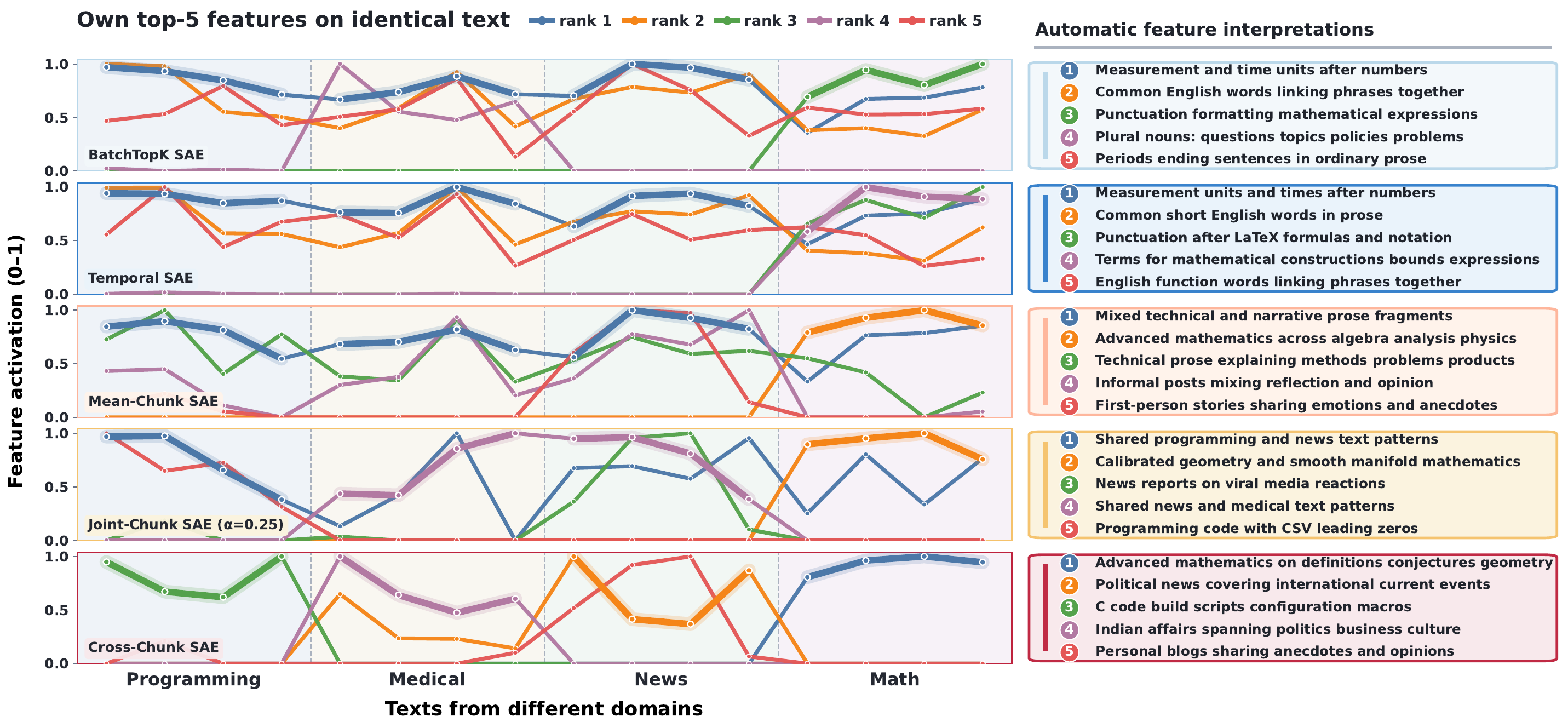}
  \vskip -.1in
  \caption{\textbf{Changes and interpretation of top-5 feature activation values for each SAE across texts from different domains.}
  Left: normalized traces of each SAE's own top-five features across programming, medical, news, and mathematics passages.
  Right: automatic interpretations matched by color. Feature identities differ across SAEs.
  \textbf{Cross-Chunk SAE exhibits sharper semantic selectivity, activating domain-relevant features while leaving unrelated features largely inactive.}}
  \label{fig:traces}
\end{figure}
\endgroup

\subsection{Fidelity: Retaining Predictable Information}
\label{sec:faithful}

Fraction of variance explained (FVE) measures error reduction
relative to a constant-mean predictor~\citep{gao2024scaling}.
Predicting a neighbor is harder than
reconstructing the observed input, so raw FVE mixes representation
quality with task difficulty. We introduce reference-normalized FVE
(RFVE), which divides each SAE's FVE by that of a reference with the
same inputs and targets:
\begin{equation}
\fve(\widehat{\mathbf{y}})
=1-\frac{\sum_i\|s\mathbf{y}_i-\widehat{\mathbf{y}}_i\|_2^2}
{\sum_i\|s(\mathbf{y}_i-\overline{\mathbf{y}})\|_2^2},
\qquad
\rfve=\frac{\fve_{\rm SAE}}{\fve_{\rm ref}}.
\label{eq:rfve}
\end{equation}
Here, $\mathbf{y}_i$ is the unscaled target,
$\overline{\mathbf{y}}$ its training-set mean, and
$\widehat{\mathbf{y}}_i$ predicts $s\mathbf{y}_i$. Self-reconstruction
uses an identity reference; partner prediction uses a train-fitted
dense predictor of width 1,024 selected on validation. Each SAE and
reference share the same evaluation monitor. Joint-Chunk weights
self and partner FVEs by $1$ and $\alpha$ in both numerator and
denominator (Appendix~\ref{app:ablations}). Target-relative fidelity provides a fair measure of SAE faithfulness within each prediction task.

\emph{Reconstruction cosine} \citep{bhalla2026temporalsae} compares the directions of a prediction and its target: their dot product divided by the product of their Euclidean norms. Higher values mean closer alignment,
independently of vector magnitude. This complements FVE, which also
penalizes errors in magnitude.

Mean-Chunk SAE reaches 93.8\% RFVE and also
leads reconstruction cosine (Figures~\ref{fig:faith_interp}
and~\ref{fig:summary}). Cross-Chunk SAE approaches its dense partner
reference despite having to predict content it never directly observes.
Its sparse code retains much of what the reference can predict.
Together, these results show chunk-level SAEs remain faithful tools for interpreting model representations.

\begingroup
\setlength{\parfillskip}{0pt plus 1fil}
\sloppy
\begin{figure}[t]
  \centering
  \includegraphics[width=1.0\linewidth]{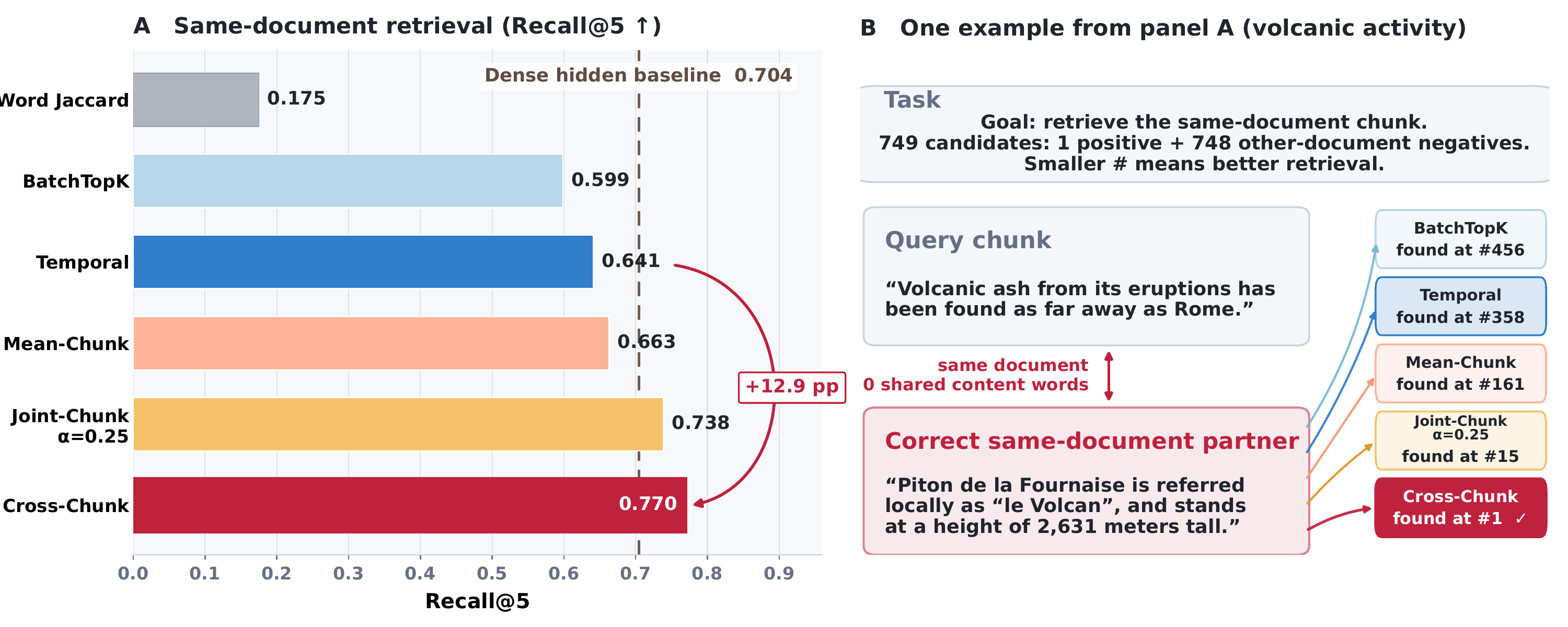}
  \vskip -.1in
\caption{\textbf{Same-document retrieval beyond lexical overlap.}
Left (A): Recall@5 for word-set Jaccard matching and SAE representations;
the dashed line marks the dense hidden-state baseline.
Right (B): an example about volcanic activity with no shared content words,
showing each SAE's rank for the correct partner among length-matched candidates.
\textbf{Chunk-level SAEs outperform token-level baselines in retrieval,
with Cross-Chunk SAE delivering substantial further gains.}}
  \label{fig:document}
\end{figure}
\endgroup

\subsection{Interpretability: Selective Explanations and High-Level Concepts}
\label{sec:interp}

\emph{Focused interpretability}, denoted InterpScore, measures the
agreement between frozen explanations and feature activation in complete
128-token contexts. We judge 1,000 feature explanations per method on
four active and four inactive contexts at temperature zero. Let $r_f^+$
be the fraction of active contexts matched and $r_f^-$ the fraction of
inactive contexts rejected. For feature $f$, we define
\begin{equation}
I_f=
\begin{cases}
\displaystyle
\frac{2r_f^+r_f^-}{r_f^++r_f^-},
& r_f^++r_f^->0,\\[6pt]
0, & r_f^++r_f^-=0,
\end{cases}
\qquad
\mathrm{InterpScore}
=\frac{100}{|\mathcal{F}|}\sum_{f\in\mathcal{F}}I_f.
\label{eq:interp}
\end{equation}
The score averages all 1,000 sampled features, including explanations
matching no active context. For all SAE families, token-level features
use whole-context activations, and chunk-level features use encoded
context means. Appendix~\ref{app:interpretability} details the full scoring
procedure.

\emph{High-level feature fraction} asks how common semantic features
are across the dictionary. We uniformly sample 1,000 coordinates and
divide the number passing a blinded semantic test by all 1,000.
A feature passes when at least six of ten strong contexts from
distinct held-out documents share a stable semantic or functional
concept, without a sufficient surface-only rule, such as a fixed
phrase or punctuation pattern (Appendix~\ref{app:prompts}). Every sampled coordinate counts,
regardless of explanation support.

\textbf{Pooling makes reusable concepts more common across the
dictionary.} High-level feature fraction reaches 34.8\% for Mean-Chunk
versus 17.2\% for BatchTopK (Figure~\ref{fig:faith_interp}).
This gain concerns what a random coordinate represents, not how
its explanation is phrased. The census complements focused
interpretability by measuring concept prevalence across all coordinates.

\subsection{Feature Structure: Persistence and Dictionary Utilization}
\label{sec:properties}

\emph{Feature persistence lift} tracks document continuity. For feature $f\in\mathcal{F}$, $\Delta_f$ is
its probability of firing in both adjacent same-document chunks
minus that probability in length-matched chunks shuffled across
documents. \emph{Dictionary utilization} measures how evenly
activation mass is spread across a sampled alive-feature set
$\mathcal{F}_{\rm alive}$ of size $N_{\rm alive}$, with $p_f$ each
feature's share of that mass~\citep{hill1973diversity}:
\begin{equation}
  \mathrm{PersistenceLift}=\frac{1}{|\mathcal{F}|}\sum_{f\in\mathcal{F}}\Delta_f,\qquad
  \mathrm{Utilization}=\frac{\exp\!\left(-\sum_{f\in\mathcal{F}_{\rm alive}}p_f\log p_f\right)}{N_{\rm alive}}.
  \label{eq:feature-properties}
\end{equation}
Persistence lift averages $\Delta_f$ across features, controlling
for frequent firing. Utilization divides the
entropy-equivalent feature count by the sampled alive-feature count:
equal shares yield 100\%, while concentration lowers it.
Together, these metrics capture feature continuity and diversity
(Appendix~\ref{app:dictionary}).

Cross-Chunk combines the strongest feature persistence lift with
32.5\% dictionary utilization, over five times BatchTopK's level
(Figure~\ref{fig:summary}). Broader dictionary use accompanies
selective passage responses: relevant features follow the content,
while unrelated features remain largely quiet
(Figure~\ref{fig:traces}). In the news passage, Cross-Chunk
concentrates activity in relevant features, while token-level traces
include responses whose explanations concern other domains.
Chunk-level SAEs thus draws on a broader range of features while selectively activating those that match each passage's content.

\begingroup
\setlength{\parfillskip}{0pt plus 1fil}
\sloppy
\begin{figure}[t]
\centering
\includegraphics[width=\linewidth]{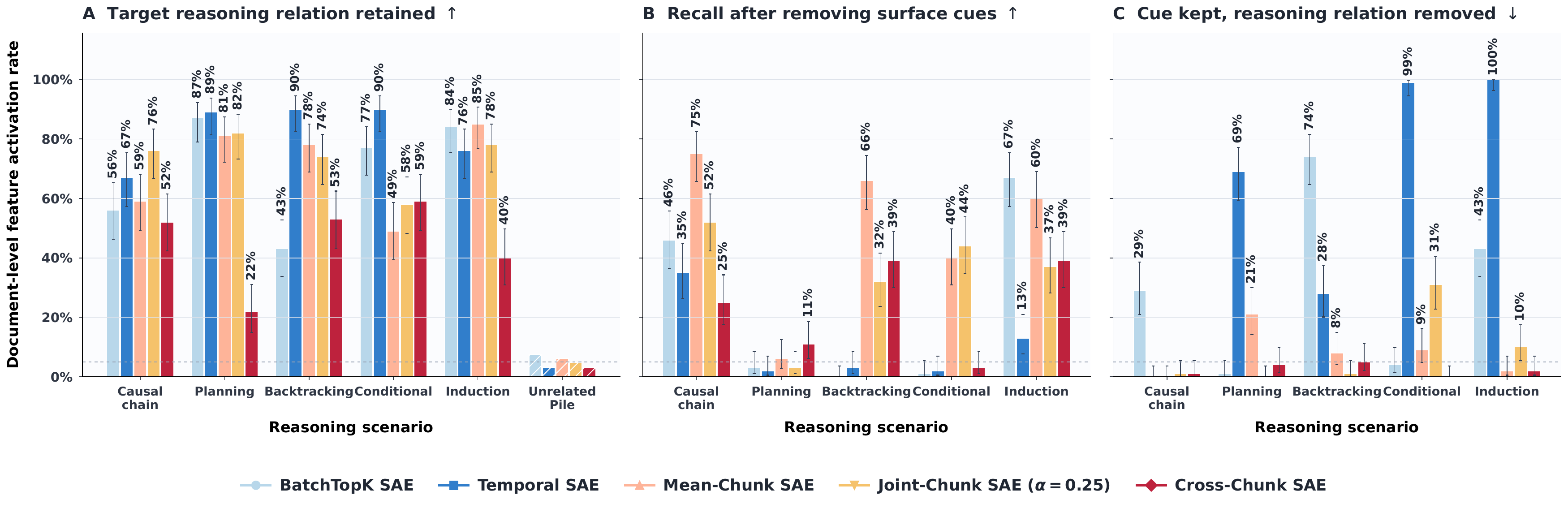}
\vskip -.1in
\caption{\textbf{Reasoning detection beyond surface cues.}
(A) Native reasoning recall on original texts with unrelated Pile controls;
(B) recall with reasoning structure preserved but surface cues removed;
(C) false feature activation with cues retained but reasoning structure removed.
\textbf{Mean-Chunk retains strong native recall, detects cue-free
reasoning, and rejects cue-only distractors.}}
\label{fig:reasoning}
\end{figure}
\endgroup

\section{Chunk-Level SAEs Improve Downstream Utility}
\label{sec:utility}

\subsection{Retrieval: Connecting Passages Beyond Shared Words}
\label{sec:document}

\emph{Document recall} tests whether SAE representations can connect
related passages that use different words. Given a query chunk, the
task is to retrieve its same-document partner from a candidate pool.
Figure~\ref{fig:summary} reports Recall@5: the fraction of queries
whose correct partner ranks among the top five candidates.
Query--partner pairs have word-set Jaccard overlap at most
0.1~\citep{manning2008ir}, and candidates match the target's token
length. We rank candidates by SAE-code cosine
similarity~\citep{kang2025retrieval}. These controls make shared
wording and passage length less useful, testing whether the codes
retain information that connects passages from the same document
(Appendix~\ref{app:protocol}).

\textbf{Cross-Chunk retrieves related content even when shared wording
offers little help.} Document recall reaches 77.0\%, compared with
64.1\% for Temporal, the strongest token baseline
(Figure~\ref{fig:document}). The volcanic-activity example makes this
ability concrete: the query and its partner share no content words,
yet their SAE codes reveal a connection that word matching misses.
Predicting neighboring chunks encourages features to capture
information shared across passages, beyond the wording of either
passage alone. This gives chunk-level saes practical value for
semantic retrieval: they can help locate related material expressed
in different terms while representing each passage through features.

\begingroup
\setlength{\parfillskip}{0pt plus 1fil}
\sloppy
\begin{figure}[t]
\centering
\includegraphics[width=\linewidth]{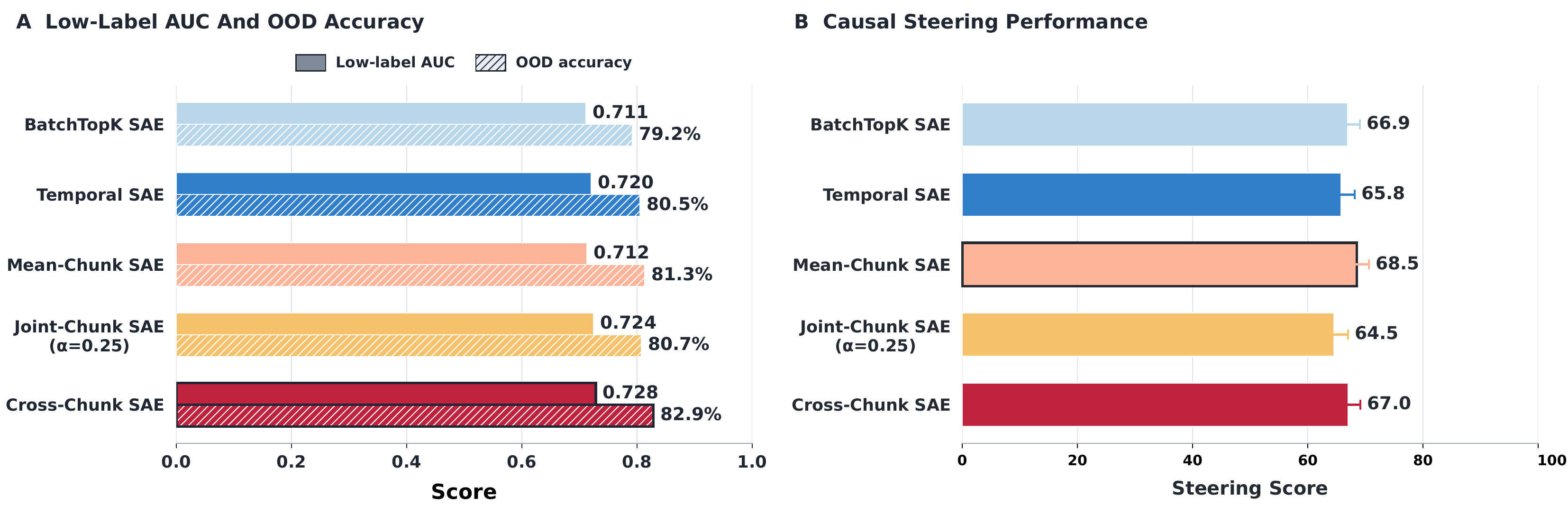}
\vskip -.1in
\caption{\textbf{Classification transfer and feature steering.}
Left (A): low-label AUC across supervision budgets (solid bars)
and OOD accuracy on future-year documents (hatched bars).
Right (B): steering scores combining feature
concept and coherence preservation; 50 denotes neutral output.
\textbf{Cross-Chunk leads both classification metrics,
while Mean-Chunk achieves the highest steering score.}}
\label{fig:transfer_steering}
\end{figure}
\endgroup

\subsection{Reasoning: Recognizing Relations Beyond Surface Cues}
\label{sec:reasoning}

We search each dictionary for causal mechanism, planning,
backtracking, conditional assumption, and induction. Rank is weighted
target activation rate minus the largest rate on other relations
or Pile background. Selection and confirmation folds require support
and specificity, with a distinct eligible feature per relation.
Cutoffs use confirmation-fold Pile background.
Feature IDs and detection cutoffs are frozen before
held-out testing and independent control rewrites
(Appendix~\ref{app:prompts}).

\emph{Native reasoning recall} is the fraction of original held-out
reasoning passages whose relation-specific feature exceeds its
fixed detection cutoff.
\emph{Reasoning generalization} is the mean of cue-free recall
and cue-only rejection. Cue-free recall measures detection
after removing verbal cues while preserving the relation; cue-only
rejection is the fraction of passages not detected when the cues
remain but the relation is removed. Together, these controls separate reasoning from familiar wording.

\textbf{These results show that chunk-level SAE features can generalize
beyond specific words to capture reasoning relations across different
wordings.} Mean-Chunk SAE achieves 66\% cue-free recall for backtracking,
compared with under 5\% for both token baselines
(Figure~\ref{fig:reasoning}). It detects the revision in a passage's
events and decisions even after words such as \emph{wait} are removed.
Cue-only controls test whether those words trigger detection when
no reasoning relation is present. Temporal SAE leads native reasoning
recall. Overall, this suggests that chunk-level SAEs, unlike token-level SAEs, do not rely heavily on the presence of a few highly activating tokens, reducing the risk of missing relevant data when such tokens are absent.

\subsection{Classification: Few Labels and Future Documents}
\label{sec:classification}

We fit multinomial logistic probes to frozen SAE representations
from eight balanced ArXiv domains, with shared regularization and
fixed data splits. \emph{OOD accuracy} is the fraction of future-year
documents assigned the correct domain. \emph{Low-label AUC}
summarizes accuracy at $\{1,2,4,8,16,64,256\}$ labels per class:
we integrate the accuracy curve over base-two log label budgets
using trapezoids and divide by the log-budget range.
Low-label results average five fixed sampling seeds
(Appendix~\ref{app:label-efficiency}).

Cross-Chunk leads both metrics, reaching 82.9\% OOD accuracy
(Figure~\ref{fig:transfer_steering}A). The two settings test different demands: exposing distinctions a probe can learn
from few examples, and preserving those distinctions as the document
collection shifts to a later period. Its retrieval advantage thus
extends to classification under limited supervision and a time shift.
Because the Pile contains ArXiv material, the OOD split does not
establish an absence of source-corpus pre-exposure
(Appendix~\ref{app:protocol}).

\subsection{Steering: Guiding Generation with Semantic Features}
\label{sec:steering}

\emph{Causal steering} tests whether a feature's decoder direction
steers generation toward its concept~\citep{wang2025utility}.
Following~\citet{wu2025axbench}, we add normalized decoder column
$\mathbf{d}_f$ to the layer-$\ell$ residual stream during prefill and
generation. A blinded judge compares the steered continuation with
the baseline, rating concept direction $C_{f,\gamma}$ and coherence
preservation $H_{f,\gamma}$ on 0--100 scales:
\begin{equation}
\mathbf{r}_{\ell}^{+}
= \mathbf{r}_{\ell}
+ \gamma\frac{\mathbf{d}_f}{\|\mathbf{d}_f\|_2},
\qquad
s_{f,\gamma}
= C_{f,\gamma}\left(0.7+0.3\frac{H_{f,\gamma}}{100}\right).
\label{eq:steering}
\end{equation}
Normalization makes $\gamma$ independent of decoder-column magnitude.
The score rewards movement toward the concept while discounting
loss of coherence; unchanged continuations score 50. We test base
strengths $\{1,3,5,7,10\}$ and adaptive rescue for eligible unchanged
outputs. For each feature, we retain the highest score among all
tested strengths, including rescue; ties favor lower strength.
The causal steering score is the mean of these maxima over
1000 features (Appendix~\ref{app:prompts} and Eq.~\ref{eq:steering-score}).

Mean-Chunk achieves the strongest causal steering performance
(Figure~\ref{fig:transfer_steering}B). Its higher score reflects more
effective steering toward the intended concept under an evaluation
that also accounts for the coherence of the generated text.
These results show that its features offer a practical way
to guide model outputs toward the semantic concepts described by
the corresponding feature explanations.

\section{Conclusion and Discussion}
\label{sec:conclusion}

We introduce a family of chunk-level SAEs that reconstruct pooled
activations (Mean-Chunk), predict independently processed neighboring
chunks (Cross-Chunk), or combine both objectives (Joint-Chunk).
\textbf{Chunk-level training retains strong target-relative fidelity
while uncovering more reliable semantic features that generalize across
wordings and respond selectively to relevant content.}
These properties give the features value beyond reconstruction.
Mean-Chunk detects reasoning without familiar cue words and achieves
the strongest causal steering, making its features useful for both
recognizing semantic patterns and guiding generation. Cross-Chunk
leads retrieval under limited lexical overlap and classification
transfer, showing that its codes capture information shared across
passages. Within this scope, learning from chunks
makes sparse dictionaries more effective tools for understanding
representations and influencing model behavior.

\par
\endgroup
\subsection*{Reproducibility Statement}
We document the SAE architectures and training objectives in
Sections~\ref{sec:tokensae} and~\ref{sec:chunksae}, and the experimental
settings and evaluation metrics in Sections~\ref{sec:faith_interp}
and~\ref{sec:utility}.
Appendix~\ref{app:protocol} provides additional training details,
evaluation splits, and representation aggregation rules.
Appendix~\ref{app:prompts} documents the judge prompts and control
procedures, while Appendix~\ref{app:ablations} reports the alpha sweep
and additional ablation results.

\subsection*{AI Use Statement}
Generative AI assisted with drafting and language editing. The authors checked all equations, numerical claims, citations, and figure captions against the training and evaluation artifacts.

\bibliography{iclr2027_conference}

@article{olshausen1996emergence,
  title = {Emergence of Simple-Cell Receptive Field Properties by Learning a Sparse Code for Natural Images},
  author = {Olshausen, Bruno A. and Field, David J.},
  journal = {Nature},
  volume = {381},
  number = {6583},
  pages = {607--609},
  year = {1996},
  doi = {10.1038/381607a0},
  url = {https://www.nature.com/articles/381607a0}
}

@article{elhage2022toy,
  title = {Toy Models of Superposition},
  author = {Elhage, Nelson and Hume, Tristan and Olsson, Catherine and Schiefer, Nicholas and Henighan, Tom and Kravec, Shauna and Hatfield-Dodds, Zac and Lasenby, Robert and Drain, Dawn and Chen, Carol and Grosse, Roger and McCandlish, Sam and Kaplan, Jared and Amodei, Dario and Wattenberg, Martin and Olah, Christopher},
  journal = {Transformer Circuits Thread},
  year = {2022},
  url = {https://transformer-circuits.pub/2022/toy_model/index.html}
}

@article{bricken2023monosemanticity,
  title = {Towards Monosemanticity: Decomposing Language Models With Dictionary Learning},
  author = {Bricken, Trenton and Templeton, Adly and Batson, Joshua and Chen, Brian and Jermyn, Adam and Conerly, Tom and Turner, Nick and Anil, Cem and Denison, Carson and Askell, Amanda and Lasenby, Robert and Wu, Yifan and Kravec, Shauna and Schiefer, Nicholas and Maxwell, Tim and Joseph, Nicholas and Hatfield-Dodds, Zac and Tamkin, Alex and Nguyen, Karina and McLean, Brayden and Burke, Josiah E and Hume, Tristan and Carter, Shan and Henighan, Tom and Olah, Christopher},
  journal = {Transformer Circuits Thread},
  year = {2023},
  url = {https://transformer-circuits.pub/2023/monosemantic-features/index.html}
}

@inproceedings{cunningham2023sparse,
  title = {Sparse Autoencoders Find Highly Interpretable Features in Language Models},
  author = {Huben, Robert and Cunningham, Hoagy and Smith, Logan and Ewart, Aidan and Sharkey, Lee},
  booktitle = {International Conference on Learning Representations},
  pages = {7827--7845},
  year = {2024},
  url = {https://proceedings.iclr.cc/paper_files/paper/2024/file/1fa1ab11f4bd5f94b2ec20e794dbfa3b-Paper-Conference.pdf}
}

@article{templeton2024scaling,
  title = {Scaling Monosemanticity: Extracting Interpretable Features from {Claude 3 Sonnet}},
  author = {Templeton, Adly and Conerly, Tom and Marcus, Jonathan and Lindsey, Jack and Bricken, Trenton and Chen, Brian and Pearce, Adam and Citro, Craig and Ameisen, Emmanuel and Jones, Andy and Cunningham, Hoagy and Turner, Nicholas L and McDougall, Callum and MacDiarmid, Monte and Freeman, C. Daniel and Sumers, Theodore R. and Rees, Edward and Batson, Joshua and Jermyn, Adam and Carter, Shan and Olah, Chris and Henighan, Tom},
  journal = {Transformer Circuits Thread},
  year = {2024},
  url = {https://transformer-circuits.pub/2024/scaling-monosemanticity/index.html}
}

@inproceedings{rajamanoharan2024gated,
  title = {Improving Sparse Decomposition of Language Model Activations with Gated Sparse Autoencoders},
  author = {Rajamanoharan, Senthooran and Conmy, Arthur and Smith, Lewis and Lieberum, Tom and Varma, Vikrant and Kram{\'a}r, J{\'a}nos and Shah, Rohin and Nanda, Neel},
  booktitle = {Advances in Neural Information Processing Systems},
  volume = {37},
  pages = {775--818},
  year = {2024},
  publisher = {Curran Associates, Inc.},
  doi = {10.52202/079017-0024},
  url = {https://proceedings.neurips.cc/paper_files/paper/2024/hash/01772a8b0420baec00c4d59fe2fbace6-Abstract-Conference.html}
}

@article{gao2024scaling,
  title = {Scaling and Evaluating Sparse Autoencoders},
  author = {Gao, Leo and Dupr{\'e} la Tour, Tom and Tillman, Henk and Goh, Gabriel and Troll, Rajan and Radford, Alec and Sutskever, Ilya and Leike, Jan and Wu, Jeffrey},
  journal = {arXiv preprint arXiv:2406.04093},
  year = {2024},
  eprint = {2406.04093},
  url = {https://arxiv.org/abs/2406.04093},
  archiveprefix = {arXiv},
  primaryclass = {cs.LG}
}

@misc{rajamanoharan2024jumprelu,
  title = {Jumping Ahead: Improving Reconstruction Fidelity with {JumpReLU} Sparse Autoencoders},
  author = {Rajamanoharan, Senthooran and Lieberum, Tom and Sonnerat, Nicolas and Conmy, Arthur and Varma, Vikrant and Kram{\'a}r, J{\'a}nos and Nanda, Neel},
  year = {2024},
  eprint = {2407.14435},
  archivePrefix = {arXiv},
  primaryClass = {cs.LG},
  doi = {10.48550/arXiv.2407.14435},
  url = {https://arxiv.org/abs/2407.14435}
}

@inproceedings{bussmann2024batchtopk,
  title = {{BatchTopK} Sparse Autoencoders},
  author = {Bussmann, Bart and Leask, Patrick and Nanda, Neel},
  booktitle = {NeurIPS 2024 Workshop on Scientific Methods for Understanding Neural Networks},
  year = {2024},
  eprint = {2412.06410},
  url = {https://neurips.cc/virtual/2024/99271},
  archiveprefix = {arXiv}
}

@inproceedings{bussmann2025matryoshka,
  title = {Learning Multi-Level Features with Matryoshka Sparse Autoencoders},
  author = {Bussmann, Bart and Nabeshima, Noa and Karvonen, Adam and Nanda, Neel},
  booktitle = {Proceedings of the 42nd International Conference on Machine Learning},
  publisher = {PMLR},
  volume = {267},
  pages = {6077--6101},
  year = {2025},
  url = {https://proceedings.mlr.press/v267/bussmann25a.html},
  series = {Proceedings of Machine Learning Research}
}

@inproceedings{bhalla2026temporalsae,
  title = {Temporal Sparse Autoencoders: Leveraging the Sequential Nature of Language for Interpretability},
  author = {Bhalla, Usha and Oesterling, Alex and Verdun, Claudio Mayrink and Lakkaraju, Himabindu and Calmon, Flavio P.},
  booktitle = {International Conference on Learning Representations},
  year = {2026},
  eprint = {2511.05541},
  url = {https://arxiv.org/abs/2511.05541},
  archiveprefix = {arXiv}
}

@article{der2026turnaveraged,
  title = {Turn-Averaged {SAEs} for Feature Discovery and Long-Context Attribution},
  author = {Der, Kevin and Kamath, Harish and Thompson, Ben},
  journal = {arXiv preprint arXiv:2606.28548},
  year = {2026},
  eprint = {2606.28548},
  url = {https://arxiv.org/abs/2606.28548},
  archiveprefix = {arXiv},
  primaryclass = {cs.CL}
}

@article{yang2026steplevel,
  title = {Step-Level Sparse Autoencoder for Reasoning Process Interpretation},
  author = {Yang, Xuan and Liu, Jiayu and Lai, Yuhang and Xu, Hao and Huang, Zhenya and Miao, Ning},
  journal = {arXiv preprint arXiv:2603.03031},
  year = {2026},
  eprint = {2603.03031},
  archivePrefix = {arXiv},
  primaryClass = {cs.LG},
  doi = {10.48550/arXiv.2603.03031},
  url = {https://arxiv.org/abs/2603.03031}
}

@inproceedings{karvonen2025saebench,
  title = {{SAEB}ench: A Comprehensive Benchmark for Sparse Autoencoders in Language Model Interpretability},
  author = {Karvonen, Adam and Rager, Can and Lin, Johnny and Tigges, Curt and Bloom, Joseph Isaac and Chanin, David and Lau, Yeu-Tong and Farrell, Eoin and Mcdougall, Callum Stuart and Ayonrinde, Kola and Till, Demian and Wearden, Matthew and Conmy, Arthur and Marks, Samuel and Nanda, Neel},
  booktitle = {Proceedings of the 42nd International Conference on Machine Learning},
  publisher = {PMLR},
  volume = {267},
  pages = {29223--29264},
  year = {2025},
  url = {https://proceedings.mlr.press/v267/karvonen25a.html},
  series = {Proceedings of Machine Learning Research}
}

@article{chanin2025sparsebutwrong,
  title = {Sparse but Wrong: Incorrect {L0} Leads to Incorrect Features in Sparse Autoencoders},
  author = {Chanin, David and Garriga-Alonso, Adri{\`a}},
  journal = {arXiv preprint arXiv:2508.16560},
  year = {2025},
  eprint = {2508.16560},
  archivePrefix = {arXiv},
  primaryClass = {cs.LG},
  doi = {10.48550/arXiv.2508.16560},
  url = {https://arxiv.org/abs/2508.16560}
}

@misc{farrell2024unlearning,
  title = {Applying Sparse Autoencoders to Unlearn Knowledge in Language Models},
  author = {Farrell, Eoin and Lau, Yeu-Tong and Conmy, Arthur},
  year = {2024},
  eprint = {2410.19278},
  archivePrefix = {arXiv},
  primaryClass = {cs.LG},
  doi = {10.48550/arXiv.2410.19278},
  url = {https://arxiv.org/abs/2410.19278}
}

@inproceedings{wang-etal-2025-model-unlearning,
    title = "Model Unlearning via Sparse Autoencoder Subspace Guided Projections",
    author = "Wang, Xu  and
      Li, Zihao  and
      Wang, Benyou  and
      Hu, Yan  and
      Zou, Difan",
    editor = "Christodoulopoulos, Christos  and
      Chakraborty, Tanmoy  and
      Rose, Carolyn  and
      Peng, Violet",
    booktitle = "Proceedings of the 2025 Conference on Empirical Methods in Natural Language Processing",
    month = nov,
    year = "2025",
    address = "Suzhou, China",
    publisher = "Association for Computational Linguistics",
    url = "https://aclanthology.org/2025.emnlp-main.1348/",
    doi = "10.18653/v1/2025.emnlp-main.1348",
    pages = "26530--26546",
    ISBN = "979-8-89176-332-6"
}

@misc{he2025saif,
  title = {{SAIF}: A Sparse Autoencoder Framework for Interpreting and Steering Instruction Following of Language Models},
  author = {He, Zirui and Zhao, Haiyan and Qiao, Yiran and Yang, Fan and Payani, Ali and Ma, Jing and Du, Mengnan},
  year = {2025},
  eprint = {2502.11356},
  archivePrefix = {arXiv},
  primaryClass = {cs.LG},
  doi = {10.48550/arXiv.2502.11356},
  url = {https://arxiv.org/abs/2502.11356}
}

@inproceedings{ma2026sparseautoencodersidentifyreasoning,
  title = {Do Sparse Autoencoders Identify Reasoning Features in Language Models?},
  author = {George Ma and Zhongyuan Liang and Irene Y. Chen and Somayeh Sojoudi},
  booktitle = {International Conference on Machine Learning},
  year = {2026},
  eprint = {2601.05679},
  url = {https://arxiv.org/abs/2601.05679},
  archiveprefix = {arXiv},
  primaryclass = {cs.LG}
}

@book{manning2008ir,
  title = {Introduction to Information Retrieval},
  author = {Manning, Christopher D. and Raghavan, Prabhakar and Sch{\"u}tze, Hinrich},
  publisher = {Cambridge University Press},
  year = {2008},
  url = {https://nlp.stanford.edu/IR-book/}
}

@inproceedings{kang2025retrieval,
  title = {Interpret and Control Dense Retrieval with Sparse Latent Features},
  author = {Hao Kang and Tevin Wang and Chenyan Xiong},
  booktitle = {Proceedings of the 2025 Conference of the Nations of the Americas Chapter of the Association for Computational Linguistics: Human Language Technologies (Volume 2: Short Papers)},
  publisher = {Association for Computational Linguistics},
  pages = {700--709},
  year = {2025},
  doi = {10.18653/v1/2025.naacl-short.58},
  url = {https://aclanthology.org/2025.naacl-short.58/}
}

@inproceedings{kantamneni2025useful,
  title = {Are Sparse Autoencoders Useful? {A} Case Study in Sparse Probing},
  author = {Kantamneni, Subhash and Engels, Joshua and Rajamanoharan, Senthooran and Tegmark, Max and Nanda, Neel},
  booktitle = {Proceedings of the 42nd International Conference on Machine Learning},
  publisher = {PMLR},
  volume = {267},
  pages = {29018--29049},
  year = {2025},
  url = {https://proceedings.mlr.press/v267/kantamneni25a.html},
  series = {Proceedings of Machine Learning Research}
}

@inproceedings{marks2024sparsefeatures,
  title = {Sparse Feature Circuits: Discovering and Editing Interpretable Causal Graphs in Language Models},
  author = {Marks, Samuel and Rager, Can and Michaud, Eric J. and Belinkov, Yonatan and Bau, David and Mueller, Aaron},
  booktitle = {International Conference on Learning Representations},
  year = {2025},
  eprint = {2403.19647},
  url = {https://arxiv.org/abs/2403.19647},
  archiveprefix = {arXiv}
}

@article{wu2025axbench,
  title = {{AxBench}: Steering {LLMs}? Even Simple Baselines Outperform Sparse Autoencoders},
  author = {Wu, Zhengxuan and Arora, Aryaman and Geiger, Atticus and Wang, Zheng and Huang, Jing and Jurafsky, Dan and Manning, Christopher D. and Potts, Christopher},
  journal = {arXiv preprint arXiv:2501.17148},
  year = {2025},
  eprint = {2501.17148},
  archivePrefix = {arXiv},
  url = {https://arxiv.org/abs/2501.17148}
}

@article{wang2025utility,
  title = {Does higher interpretability imply better utility? A Pairwise Analysis on Sparse Autoencoders},
  author = {Wang, Xu and Hu, Yan and Wang, Benyou and Zou, Difan},
  journal = {arXiv preprint arXiv:2510.03659},
  year = {2025},
  eprint = {2510.03659},
  archivePrefix = {arXiv},
  url = {https://arxiv.org/abs/2510.03659}
}

@article{oord2018cpc,
  title = {Representation Learning with Contrastive Predictive Coding},
  author = {van den Oord, Aaron and Li, Yazhe and Vinyals, Oriol},
  journal = {arXiv preprint arXiv:1807.03748},
  year = {2018},
  eprint = {1807.03748},
  archivePrefix = {arXiv},
  url = {https://arxiv.org/abs/1807.03748}
}

@misc{qwen2026qwen35,
  title = {{Qwen3.5}: Towards Native Multimodal Agents},
  author = {{Qwen Team}},
  year = {2026},
  month = {February},
  url = {https://qwen.ai/blog?id=qwen3.5}
}

@misc{qwen2026qwen35base,
  title = {{Qwen3.5-9B-Base}},
  author = {{Qwen Team}},
  year = {2026},
  url = {https://huggingface.co/Qwen/Qwen3.5-9B-Base},
  howpublished = {Hugging Face model card},
  note = {Accessed September 9, 2026}
}

@article{deepseek2026v4,
  title = {{DeepSeek-V4}: Towards Highly Efficient Million-Token Context Intelligence},
  author = {{DeepSeek-AI} and others},
  journal = {arXiv preprint arXiv:2606.19348},
  year = {2026},
  eprint = {2606.19348},
  archivePrefix = {arXiv},
  primaryClass = {cs.CL},
  url = {https://arxiv.org/abs/2606.19348}
}

@article{gao2020pile,
  title = {The {Pile}: An {800GB} Dataset of Diverse Text for Language Modeling},
  author = {Gao, Leo and Biderman, Stella and Black, Sid and Golding, Laurence and Hoppe, Travis and Foster, Charles and Phang, Jason and He, Horace and Thite, Anish and Nabeshima, Noa and Presser, Shawn and Leahy, Connor},
  journal = {arXiv preprint arXiv:2101.00027},
  year = {2020},
  eprint = {2101.00027},
  url = {https://arxiv.org/abs/2101.00027},
  archiveprefix = {arXiv}
}

@inproceedings{lee2022deduplicating,
  title = {Deduplicating Training Data Makes Language Models Better},
  author = {Katherine Lee and Daphne Ippolito and Andrew Nystrom and Chiyuan Zhang and Douglas Eck and Chris Callison-Burch and Nicholas Carlini},
  booktitle = {Proceedings of the 60th Annual Meeting of the Association for Computational Linguistics (Volume 1: Long Papers)},
  publisher = {Association for Computational Linguistics},
  pages = {8424--8445},
  year = {2022},
  doi = {10.18653/v1/2022.acl-long.577},
  url = {https://aclanthology.org/2022.acl-long.577/}
}

@misc{arxiv2026taxonomy,
  title = {{arXiv} Category Taxonomy},
  author = {{arXiv}},
  year = {n.d.},
  url = {https://arxiv.org/category_taxonomy},
  howpublished = {Official subject classification}
}

@inproceedings{kingma2015adam,
  title = {{Adam}: A Method for Stochastic Optimization},
  author = {Kingma, Diederik P. and Ba, Jimmy},
  booktitle = {International Conference on Learning Representations},
  year = {2015},
  eprint = {1412.6980},
  archivePrefix = {arXiv},
  url = {https://arxiv.org/abs/1412.6980}
}

@inproceedings{he2016residual,
  title = {Deep Residual Learning for Image Recognition},
  author = {He, Kaiming and Zhang, Xiangyu and Ren, Shaoqing and Sun, Jian},
  booktitle = {Proceedings of the IEEE Conference on Computer Vision and Pattern Recognition},
  pages = {770--778},
  year = {2016},
  eprint = {1512.03385},
  archivePrefix = {arXiv},
  url = {https://arxiv.org/abs/1512.03385}
}

@article{hill1973diversity,
  title = {Diversity and Evenness: A Unifying Notation and Its Consequences},
  author = {Hill, M. O.},
  journal = {Ecology},
  volume = {54},
  number = {2},
  pages = {427--432},
  year = {1973},
  doi = {10.2307/1934352},
  url = {https://esajournals.onlinelibrary.wiley.com/doi/10.2307/1934352}
}

@article{efron1979bootstrap,
  title = {Bootstrap Methods: Another Look at the Jackknife},
  author = {Efron, Bradley},
  journal = {The Annals of Statistics},
  volume = {7},
  number = {1},
  pages = {1--26},
  year = {1979},
  doi = {10.1214/aos/1176344552},
  url = {https://doi.org/10.1214/aos/1176344552}
}

@article{wilson1927probable,
  title = {Probable Inference, the Law of Succession, and Statistical Inference},
  author = {Wilson, Edwin B.},
  journal = {Journal of the American Statistical Association},
  volume = {22},
  number = {158},
  pages = {209--212},
  year = {1927},
  doi = {10.1080/01621459.1927.10502953},
  url = {https://doi.org/10.1080/01621459.1927.10502953}
}

@misc{ba2016layernorm,
  title = {Layer Normalization},
  author = {Ba, Jimmy Lei and Kiros, Jamie Ryan and Hinton, Geoffrey E.},
  year = {2016},
  eprint = {1607.06450},
  archivePrefix = {arXiv},
  primaryClass = {stat.ML},
  url = {https://arxiv.org/abs/1607.06450}
}

@article{elfwing2018silu,
  title = {Sigmoid-Weighted Linear Units for Neural Network Function Approximation in Reinforcement Learning},
  author = {Elfwing, Stefan and Uchibe, Eiji and Doya, Kenji},
  journal = {Neural Networks},
  volume = {107},
  pages = {3--11},
  year = {2018},
  doi = {10.1016/j.neunet.2017.12.012},
  url = {https://arxiv.org/abs/1702.03118}
}

@article{vandermaaten2008tsne,
  title = {Visualizing Data using {t-SNE}},
  author = {van der Maaten, Laurens and Hinton, Geoffrey},
  journal = {Journal of Machine Learning Research},
  volume = {9},
  number = {86},
  pages = {2579--2605},
  year = {2008},
  url = {https://jmlr.org/papers/v9/vandermaaten08a.html}
}

@article{rousseeuw1987silhouettes,
  title = {Silhouettes: A Graphical Aid to the Interpretation and Validation of Cluster Analysis},
  author = {Rousseeuw, Peter J.},
  journal = {Journal of Computational and Applied Mathematics},
  volume = {20},
  pages = {53--65},
  year = {1987},
  doi = {10.1016/0377-0427(87)90125-7},
  url = {https://doi.org/10.1016/0377-0427(87)90125-7}
}
\bibliographystyle{iclr2027_conference}

\appendix
\clearpage
\begingroup
\raggedbottom
\allowdisplaybreaks[1]
\setlength{\parfillskip}{0pt plus 1fil}
\tolerance=1500
\emergencystretch=0pt
\clubpenalty=10000
\widowpenalty=10000
\displaywidowpenalty=10000
\setcounter{topnumber}{1}
\setcounter{totalnumber}{5}
\setcounter{bottomnumber}{2}
\setlength{\textfloatsep}{14pt}
\setlength{\intextsep}{10pt}
\makeatletter
\setlength{\@fptop}{0pt}
\setlength{\@fpbot}{0pt plus 1fil}
\setlength{\@fpsep}{\textheight}
\makeatother

\section{Training and Evaluation Protocols}
\label{app:protocol}

Our comparison varies the observation unit and prediction target while sharing the model, token stream, dictionary width, and sparse-coding backbone. This appendix specifies how the five SAE families are trained and how their representations enter each evaluation. The descriptions complement the objectives in Sections~\ref{sec:tokensae} and~\ref{sec:chunksae}; judge instructions and metric definitions appear in Appendices~\ref{app:prompts} and~\ref{app:metric-definitions}. Notation and evaluation units follow the corresponding main-text experiment.

\subsection{Model, corpus, and training schedule}

We extract layer-21 activations from Qwen3.5-9B-Base~\citep{qwen2026qwen35base}, with hidden width $d=4096$. Every dictionary contains $m=65{,}536$ coordinates and uses BatchTopK budget $K=128$~\citep{bussmann2024batchtopk}. The common Pile stream~\citep{gao2020pile} contains exactly one billion training token occurrences. Validation and test caches are disjoint, with $10{,}000{,}128$ occurrences each. Activations are cached in bfloat16; dictionary parameters and optimizer accumulators use float32. Cached states and learned dictionaries thus use distinct numerical precision.

Training comprises $31{,}250$ updates with a global batch of $32{,}000$ token occurrences, distributed across eight ranks. Adam~\citep{kingma2015adam} uses $(\beta_1,\beta_2)=(0.9,0.999)$. The learning rate reaches $10^{-4}$ after 200 warmup updates and follows cosine decay to $10^{-5}$; gradients are clipped at 1.0. All methods share the activation scale, decoder-column normalization, data order, and validation-based checkpoint-selection procedure. All main runs use this schedule.

The AuxK coefficient is $\lambda_{\rm aux}=0.0625$~\citep{gao2024scaling}. A feature becomes eligible for rescue after five million inactive occurrences, while the training-health diagnostic tracks inactivity over ten million occurrences. These settings retain a common feature-rescue mechanism across the five dictionary families. Rescue accompanies the reconstruction terms defined in the main text.

Adjacent chunks $A$ and $B$ are non-overlapping spans from the same document. Their lengths belong to $\{32,64,128,256,512\}$, covering all 25 ordered length combinations. Each chunk receives a separate model forward with position indices reset, so its hidden states contain no attention to its partner. The resulting caches contain 2,520,159 training pairs, 25,202 validation pairs, and 25,201 test pairs, as summarized in Table~\ref{tab:protocol}. Pair and token counts describe the same cached streams.

\begin{table}[ht]
\centering
\small
\caption{Shared model, training budget, and chunk-pair caches for all five SAE families and every Joint-Chunk partner weight. Token occurrences account for data volume; each objective retains its own encoding and loss-averaging unit for the token-level or chunk-level prediction task.}
\label{tab:protocol}
\begin{tabular*}{\linewidth}{@{\extracolsep{\fill}}lr@{}}
\toprule
Quantity & Value \\
\midrule
Model layer / hidden width & 21 / 4096 \\
Dictionary width / BatchTopK budget & 65,536 / 128 \\
Training token occurrences & 1,000,000,000 \\
Validation / test occurrences & 10,000,128 / 10,000,128 \\
Training updates / global token batch & 31,250 / 32,000 \\
Chunk lengths & 32, 64, 128, 256, 512 \\
Training / validation / test pairs & 2,520,159 / 25,202 / 25,201 \\
\bottomrule
\end{tabular*}
\end{table}

Token baselines average reconstruction error over valid token positions. Mean-Chunk averages over sampled chunks, with one loss per chunk regardless of length. Cross-Chunk equally weights the two directed predictions in a pair. Joint-Chunk averages the directed self/partner objective in Eq.~\ref{eq:joint-task}. This distinguishes training-text volume from the unit represented by each individual sparse code.

\subsection{Representations, targets, and Joint-Chunk normalization}

Let $\psi_M$ denote the inference mapping for method $M$, consisting of its encoder followed by its fixed inference-time thresholding rule. For a chunk $C$, let $V(C)$ be the set of valid, non-padding token positions and let $L=|V(C)|$. If $\mathbf{h}_t$ denotes the representation at position $t$ within the chunk, the chunk mean is $\boldsymbol{\mu}_C=L^{-1}\sum_{t\in V(C)}\mathbf{h}_t$. We group the methods into token-level methods, $\mathcal{M}_{\mathrm{token}}=\{\text{BatchTopK},\text{Temporal}\}$, and chunk-level methods, $\mathcal{M}_{\mathrm{chunk}}=\{\text{Mean-Chunk},\text{Cross-Chunk},\text{Joint-Chunk}\}$. Retrieval and classification use the representations defined below. Whenever token representations are averaged, padding positions are excluded from both the sum and the token count, so the average includes only valid tokens in the relevant input:
\begin{equation}
\mathbf{z}_M(C)=
\begin{cases}
\displaystyle L^{-1}\sum_{t\in C}\psi_M(\mathbf{h}_t),
&M\in\mathcal{M}_{\rm token},\\[4pt]
\psi_M(\boldsymbol{\mu}_C),
&M\in\mathcal{M}_{\rm chunk}.
\end{cases}
\label{eq:appendix-representations}
\end{equation}
Token methods therefore threshold before pooling, whereas chunk methods pool before encoding and thresholding. Each representation depends only on the observed chunk. The neighboring hidden states supply a prediction target during training and fidelity evaluation, while retrieval and classification compare the observed chunks' own codes. These codes support both ranking and classification.

Joint-Chunk produces one full code
$\mathbf{z}_A\in\mathbb{R}^{m}$ with $m=65{,}536$.
Its self decoder reads the full code, whereas its partner decoder
reads the first $h=32{,}768$ coordinates. Both predictions use the
same decoder columns for this prefix, with separate self and partner
biases as in Eq.~\ref{eq:joint-decode}. The prefix is sliced after
the single BatchTopK operation; it is not re-encoded or assigned
another main-task sparse budget.

For target type $q\in\{\mathrm{self},\mathrm{partner}\}$, let
$\mathcal{T}_q$ be its training target set,
$N_q=|\mathcal{T}_q|$, and
$\overline{\mathbf{y}}^{(q)}_{\rm train}$ its mean. We normalize
each per-example squared Euclidean error by the corresponding
\emph{per-target average} constant-predictor error:
\begin{equation}
b_q
=\frac{1}{N_q}
\sum_{i\in\mathcal{T}_q}
\left\|
s\left(
\mathbf{y}^{(q)}_i-
\overline{\mathbf{y}}^{(q)}_{\rm train}
\right)
\right\|_2^2.
\label{eq:joint-normalizers}
\end{equation}
Self targets are observed chunk means, and partner targets are
neighboring chunk means. The constants $b_{\rm self}$ and
$b_{\rm partner}$ are computed once from training targets and remain
fixed during optimization. Equation~\ref{eq:joint-task} divides each
directed example's self and partner reconstruction errors by these
respective constants.

The AuxK term uses the same target normalization. Let
$\mathcal{D}_{\rm dead}$ be the coordinates eligible for inactive-feature
rescue, and let $\mathbf{u}^{\rm aux}_A$ retain the largest
$K_{\rm aux}$ positive encoder activations among those coordinates,
setting all other coordinates to zero. This auxiliary code is used
only in the rescue loss. Define the main-prediction residuals and
their auxiliary reconstructions by
\begin{equation}
\begin{aligned}
\mathbf{e}_{\rm self}(A)
&=s\boldsymbol{\mu}_A
  -\widehat{\mathbf{y}}_{\rm self}(\boldsymbol{\mu}_A),
&
\widehat{\mathbf{e}}_{\rm self}^{\rm aux}(A)
&=W_{\rm dec}\mathbf{u}^{\rm aux}_A,\\
\mathbf{e}_{\rm partner}(A,B)
&=s\boldsymbol{\mu}_B
  -\widehat{\mathbf{y}}_{\rm partner}(\boldsymbol{\mu}_A),
&
\widehat{\mathbf{e}}_{\rm partner}^{\rm aux}(A)
&=W_{\rm dec}[:,1:h]\,
  \mathbf{u}^{\rm aux}_{A,1:h}.
\end{aligned}
\label{eq:joint-aux-residuals}
\end{equation}
For a directed pair, the normalized auxiliary loss is
\begin{equation}
\begin{aligned}
\ell_{\rm AuxK}(A,B;\alpha)
=\frac{1}{1+\alpha}\Bigg[
&\frac{
\left\|\mathbf{e}_{\rm self}(A)
-\widehat{\mathbf{e}}_{\rm self}^{\rm aux}(A)\right\|_2^2
}{b_{\rm self}}\\
&+\alpha
\frac{
\left\|\mathbf{e}_{\rm partner}(A,B)
-\widehat{\mathbf{e}}_{\rm partner}^{\rm aux}(A)\right\|_2^2
}{b_{\rm partner}}
\Bigg],
\end{aligned}
\label{eq:joint-aux-per-pair}
\end{equation}
and
$\mathcal{L}_{\rm AuxK}
=\mathbb{E}_{(A,B)}[\ell_{\rm AuxK}(A,B;\alpha)]$.
Thus, both the main loss and AuxK use minibatch means, the same
self-to-partner weighting $1:\alpha$, and the same respective
per-target scales $b_{\rm self}$ and $b_{\rm partner}$. Their
relative coefficient in Eq.~\ref{eq:joint} is therefore
$\lambda_{\rm aux}=0.0625$, independently of the number of
training pairs. At $\alpha=0.25$, the normalized self and partner
components each receive weights $0.8$ and $0.2$, respectively.

\subsection{Token-level baseline objectives}
\label{app:baseline-objectives}

BatchTopK reconstructs contextualized token states with the scaled error from Section~\ref{sec:tokensae}. Averaging over valid positions and adding the shared AuxK term gives the token-level objective:
\begin{equation}
\mathcal{L}_{\rm token}
=\mathbb{E}_t\mathcal{L}_{\rm rec}(\mathbf{h}_t,\mathbf{h}_t)
+\lambda_{\rm aux}\mathcal{L}_{\rm AuxK}.
\label{eq:token}
\end{equation}
Temporal~\citep{bhalla2026temporalsae} also reconstructs token states and couples adjacent-token prefix codes. Its designated prefix has width $h_T=\lfloor0.2m\rfloor=13{,}107$, giving $\mathbf{z}^{(H)}_t=\mathbf{z}(\mathbf{h}_t)[1:h_T]$. This token-level prefix and Joint-Chunk's 32,768-coordinate shared prefix are defined separately for their respective objectives. Each prefix serves its own objective's information-sharing task.

Let $\mathcal{B}$ contain valid current tokens in a minibatch, and let $\mathcal{P}\subseteq\mathcal{B}$ contain those with an immediate predecessor in the same independently processed chunk. The full-code term includes chunk-first tokens; the prefix term averages over anchors with a predecessor. The objective combines full-code and prefix reconstruction, contrastive alignment, and the shared inactive-feature rescue term
\begin{equation}
\begin{aligned}
\mathcal{L}_{\rm full}
&=\mathbb{E}_{t\in\mathcal{B}}
\mathcal{L}_{\rm rec}(\mathbf{h}_t,\mathbf{h}_t),\\
\mathcal{L}_{H}
&=\mathbb{E}_{t\in\mathcal{P}}
\left\|W_{\rm dec}[:,1:h_T]\mathbf{z}^{(H)}_t
+\mathbf{b}_{\rm out}-s\mathbf{h}_t\right\|_2^2,\\
\mathcal{L}_{\rm temporal}
&=\frac{0.8\mathcal{L}_{\rm full}+0.2\mathcal{L}_{H}}{2}
+\mathcal{L}_{\rm InfoNCE}^{\rm sym}
+0.0625\mathcal{L}_{\rm AuxK}.
\end{aligned}
\label{eq:temporal}
\end{equation}
For the $N$ valid adjacent pairs gathered across ranks, let $\mathbf{u}_i$ and $\mathbf{v}_i$ be current- and previous-token prefix codes. With cosine similarity, temperature $\tau=0.1$, and $A_{ij}=\operatorname{sim}(\mathbf{u}_i,\mathbf{v}_j)/\tau$, symmetric InfoNCE~\citep{oord2018cpc} averages the two matching directions over all paired anchors:
\begin{equation}
\mathcal{L}_{\rm InfoNCE}^{\rm sym}
=-\frac{1}{2N}\sum_{i=1}^{N}
\left[
\log\frac{\exp(A_{ii})}{\sum_{j=1}^{N}\exp(A_{ij})}
+\log\frac{\exp(A_{ii})}{\sum_{j=1}^{N}\exp(A_{ji})}
\right].
\label{eq:infonce}
\end{equation}
Diagonal entries identify immediate within-chunk positives; off-diagonal entries supply negatives without an additional token-identity exclusion. Temporal thus aligns neighboring token codes while preserving token reconstruction. Cross-Chunk changes the prediction target to the mean activation of an independently processed neighboring passage, targeting shared passage content.

\subsection{Fidelity references and statistical summaries}

Each fidelity reference uses the same observed inputs, targets, and evaluation examples as its SAE. Self-reconstruction uses the identity reference, whose FVE equals one. Partner prediction uses a dense residual MLP with LayerNorm, a linear skip, and a SiLU residual branch~\citep{ba2016layernorm,he2016residual,elfwing2018silu}. The predictor receives the observed chunk mean without a prediction-direction flag. Both directions use this same dense reference architecture and parameters.

Reference widths $\{1024,2048,4096\}$ are trained for two epochs. Selecting the smallest width within 0.002 FVE of the best validation score yields width 1,024. Its FVE is 0.676758 on the fixed 262,144-occurrence monitor. Each SAE score is divided by the reference score from its corresponding evaluation set, preserving the same examples and target throughout the RFVE calculation. This alignment also applies to the two separately evaluated prediction targets. But using FVE for all SAEs does not constitute an apples-to-apples comparison. A truly fair comparison involves measuring performance within their respective tasks. True fairness requires ensuring that each SAE serves as a faithful tool relative to its specific objective. Our design precisely embodies this fairness.

Intervals use 10,000 paired bootstrap resamples unless a panel specifies another procedure~\citep{efron1979bootstrap}. Retrieval uses 100,000 paired randomization permutations. The high-level census reports Wilson 95\% intervals~\citep{wilson1927probable}, and the matched reasoning audit uses 20,000 paired bootstrap resamples. These summaries quantify variation over the evaluated examples or features for the checkpoints used in each comparison. Resampling pairs the evaluated examples or features.

\subsection{Retrieval and classification data}

Each query-specific retrieval gallery contains 749 candidates: one same-document partner and 748 distractor chunks from other documents, each matched to the partner's length. Each query is paired with a different chunk from its document, subject to Jaccard overlap at most 0.10. The gallery contains the partner and distractors from other documents. Every method ranks the same query-specific gallery: the lexical baseline uses Jaccard similarity, the dense baseline uses hidden-state representations, and SAEs use cosine similarity between the codes in Eq.~\ref{eq:appendix-representations}. This production-critical setting calls for better SAEs: token-level methods fall short, while our approach closes the gap.

Classification uses eight balanced ArXiv domains: Computer Science, Economics, Electrical Engineering, Mathematics, Physics, Quantitative Biology, Quantitative Finance, and Statistics~\citep{arxiv2026taxonomy}. Each class has 1,024 training, 256 validation, and 256 test documents. The OOD collection contains papers first submitted in 2023 or later, fixed before feature extraction. Frozen-code multinomial logistic probes share $C=1$, L-BFGS, and a 2,000-iteration limit for every method.

The ArXiv feature archive retains native tokenized lengths up to 512 and excludes padding through the attention mask. Low-label budgets are $\{1,2,4,8,16,64,256\}$ documents per class, with five sampling seeds shared across methods. The publication-year split evaluates transfer from the probe's training distribution to later documents. The geometry analysis uses the same document codes with the label-free dimensionality reduction described in Appendix~\ref{app:geometry}. These neighborhoods connect scientific domains to the local arrangement of frozen sparse document representations.

\section{Training Stability and Feature Activity}
\label{app:training-health}

\begin{figure}[t]
\centering
\includegraphics[width=\linewidth]{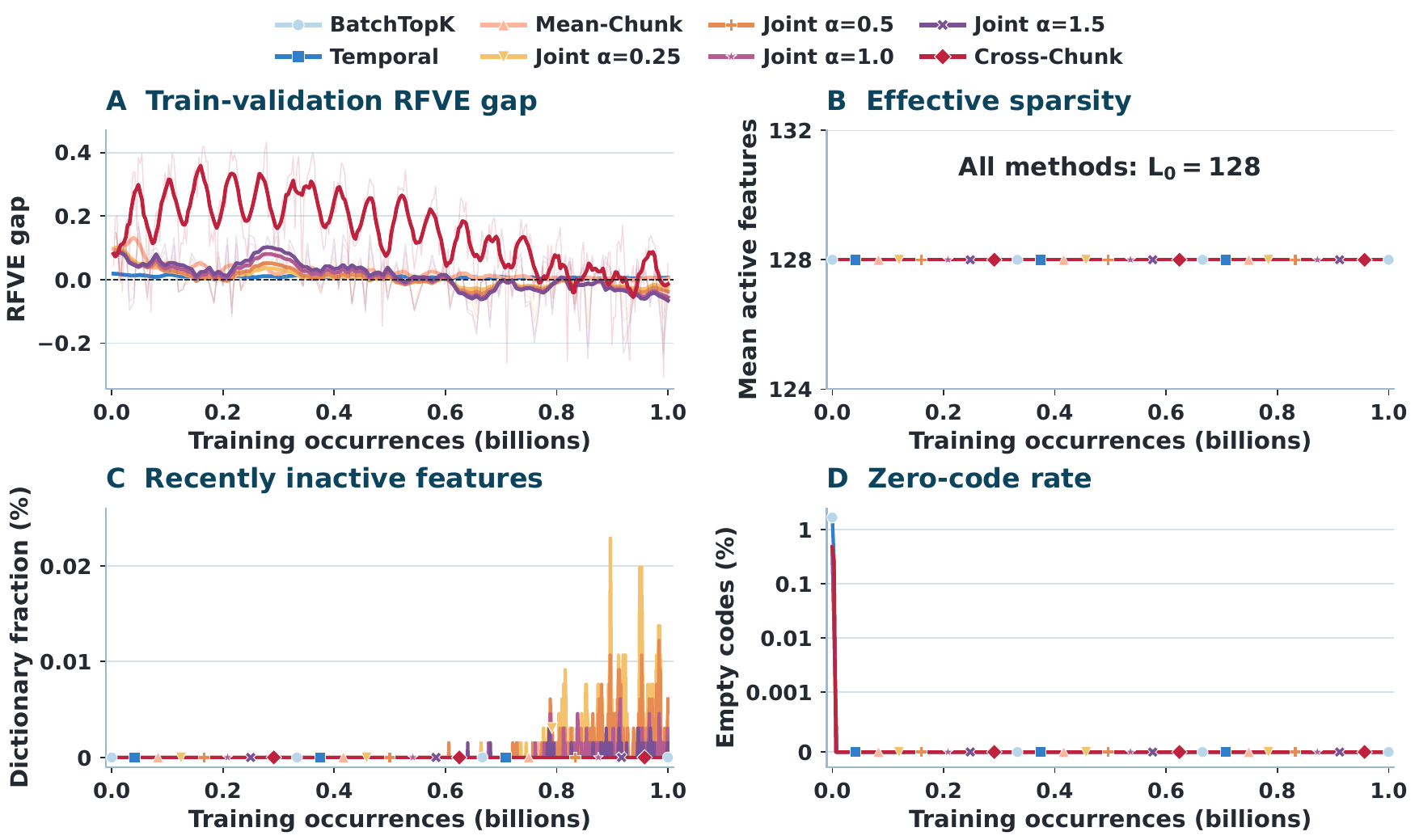}
\caption{\textbf{Training stability and sparse-code activity.}
(A) Training-batch and fixed-monitor RFVE, shown as logged values and
nine-observation moving averages.
(B) Validation effective $L_0$, the mean number of active coordinates
per code. (the sparsity variation curve demonstrates stability.)
(C) Fraction of dictionary coordinates that have not activated in the
preceding ten million token occurrences.
(D) Validation empty-code rate; the symmetric-log axis has a linear
region below 0.001\%.
Coincident curves use staggered markers. For Joint-Chunk, the activity
summaries show the smaller effective $L_0$ and larger empty-code rate
across its full-code and partner-prefix views.}
\label{fig:training-health-appendix}
\end{figure}

Figure~\ref{fig:training-health-appendix} follows reconstruction
fidelity and feature activity throughout training. Panel A compares
RFVE on the current training batch with RFVE on a fixed
262,144-occurrence validation monitor. The faint curves retain
individual observations; the nine-observation moving averages make
longer-term changes easier to inspect. Each validation RFVE uses a
reference evaluated on that same monitor.

The activity panels examine different parts of the sparse code.
Effective $L_0$ counts nonzero coordinates per evaluated input, while
the empty-code rate records inputs for which no coordinate activates.
Recent inactivity instead asks whether each dictionary coordinate
has activated at least once during the preceding ten million token
occurrences. Together, these measurements distinguish activity
within individual codes from continued use of the dictionary across
the data stream.

At the selected checkpoints, the validation empty-code rate is zero.
Once the ten-million-occurrence window is available, fewer than
0.025\% of dictionary coordinates are recently inactive at any
logged step. These observations show that the reported checkpoints
produce nonempty sparse codes while continuing to use nearly all
dictionary coordinates.

The budget $K=128$ applies to Joint-Chunk's full code, not
separately to its partner prefix. Because the partner decoder reads
the first $h$ coordinates of that code, their realized activities
satisfy
\begin{equation}
\|\mathbf{z}\|_0
=
\|\mathbf{z}_{1:h}\|_0
+
\|\mathbf{z}_{h+1:m}\|_0.
\label{eq:joint-prefix-activity}
\end{equation}
The identity separates full-code activity from the activity
available to partner prediction. Both predictions use coordinates
selected by the same encoder pass; partner decoding introduces no
additional feature-selection step. The fidelity and activity traces
therefore characterize the dictionaries and sparse codes used by
the reported evaluations.

\section{Focused Interpretability and High-Level Features}
\label{app:interpretability}

\subsection{Focused Interpretability}

For each SAE family, we evaluate 1,000 frozen feature explanations.
The judge examines four active and four inactive complete 128-token
contexts per feature at temperature zero. Token-level features are
evaluated through whole-context activations, and chunk-level features
through encoded context means.

Let $a_f$ be the number of active contexts matched by the explanation,
and let $b_f$ be the number of inactive contexts it correctly rejects.
Define $r_f^+=a_f/4$ and $r_f^-=b_f/4$. The feature score and its
dictionary-level average are
\begin{equation}
I_f=
\begin{cases}
\displaystyle
\frac{2r_f^+r_f^-}{r_f^++r_f^-},
& r_f^++r_f^->0,\\[6pt]
0, & r_f^++r_f^-=0,
\end{cases}
\qquad
\mathrm{InterpScore}
=\frac{100}{|\mathcal{F}|}
\sum_{f\in\mathcal{F}}I_f,
\label{eq:focused-interpretability}
\end{equation}
where $\mathcal{F}$ contains all 1,000 sampled features. A feature
receives zero when either active-context coverage or inactive-context
rejection is zero. InterpScore averages the feature scores and is
reported on a 0--100 scale.

The high-level feature census in the following subsection uses its
separate ten-context criterion and does not enter InterpScore.

\subsection{High-level feature census}

\looseness=-1 The census uniformly samples 1,000 coordinates from each complete dictionary. A blinded judge sees ten strong contexts from distinct held-out documents, without method names or activation magnitudes. A coordinate passes when a stable semantic or functional concept explains at least six examples and no sufficient surface-only rule accounts for the pattern. Appendix~\ref{app:prompts} reproduces the rubric used to distinguish semantic content from fixed phrases, formatting, and other surface regularities.

All sampled coordinates remain in the denominator. Mean-Chunk's 34.8\% and BatchTopK's 17.2\% therefore correspond to 348 and 172 passing coordinates, respectively, with Wilson 95\% intervals shown in Figure~\ref{fig:faith_interp}. Focused interpretability measures how consistently a frozen explanation
agrees with a feature's activation pattern across active and inactive
contexts. Together, they connect explanation quality with reusable concepts across the learned sparse representation.

\section{Dictionary Utilization and Feature Dynamics}
\label{app:dictionary}

The dictionary audit uses 2,048 validation pairs and 8,192 sampled coordinates, with a minimum-support threshold of eight. Feature persistence lift compares coactivation in adjacent same-document chunks with coactivation in length-matched chunks shuffled across documents. Shuffling preserves the lengths of the input units while changing their document relationship, allowing the audit to identify continuity associated with related content. Both pair collections use the same feature-activation rule when estimating their coactivation probabilities over pairs of independently processed passages.

Dictionary utilization summarizes how activation mass is distributed within the sampled alive-feature set. We sum each feature's nonnegative activations over the audit contexts, normalize these masses, and compute the entropy-equivalent feature count~\citep{hill1973diversity}. Dividing by the sampled alive-feature count yields Eq.~\ref{eq:utilization}. Equal shares produce 100\% utilization; concentration on a smaller subset produces a lower value, even when many coordinates have fired. The normalization expresses this distribution relative to the number of alive coordinates retained in the sampled dictionary audit.

Figure~\ref{fig:traces} illustrates local behavior with each SAE's own five selected features and their frozen explanations. The colors connect each explanation to its trace, making it possible to inspect where a described concept becomes active as the passage changes domain. The held-out persistence and utilization audit complements these examples with aggregate measurements of continuity and activation-mass distribution across the sampled dictionary. Both connect features to content.

\section{Semantic Neighborhoods Across Domains and Time}
\label{app:geometry}

\begin{figure}[t]
\centering
\includegraphics[width=\linewidth]{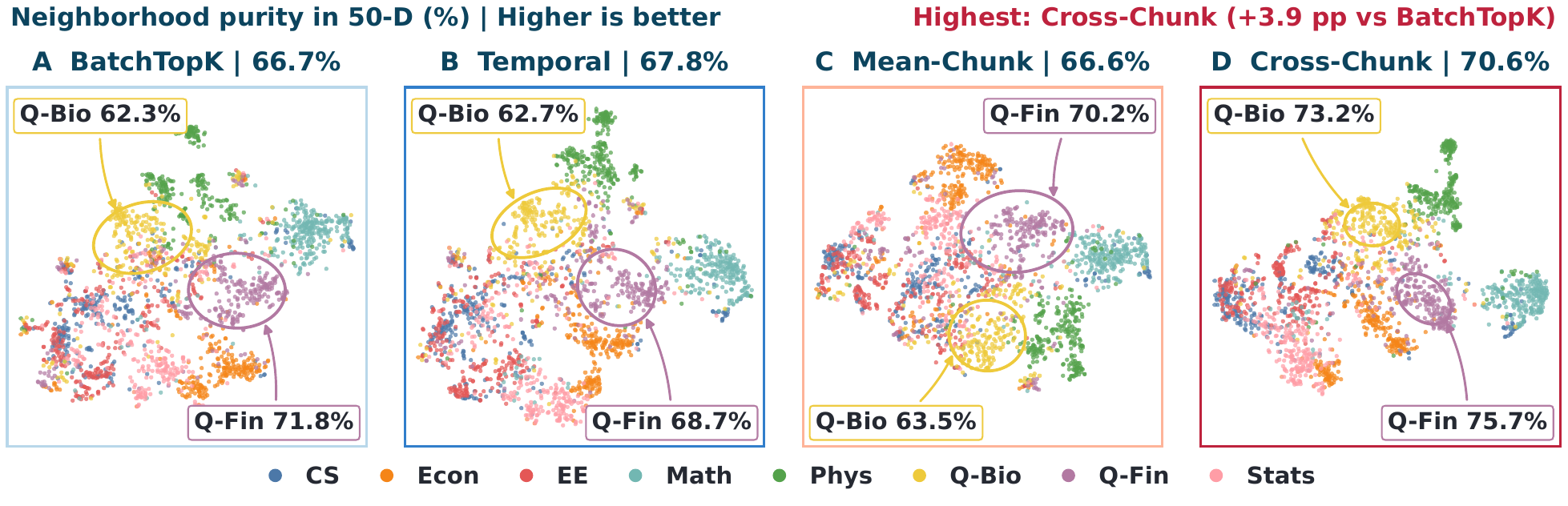}
\caption{\textbf{Scientific-domain structure in document representations.} Independently fitted t-SNE projections of the same 2,048 held-out ArXiv document codes for BatchTopK, Temporal, Mean-Chunk, and Cross-Chunk. Colors identify domains. Titles and callouts report purity in the cached 50-dimensional cosine neighborhoods, using non-self ranks 2--11. Cross-Chunk has the highest overall purity among the displayed methods, with domain-specific values highlighted in the callouts.}
\label{fig:document-neighborhoods-appendix}
\end{figure}

The geometry audit examines the organization of 2,048 held-out ArXiv documents balanced across eight domains. Frozen codes are projected to 50 dimensions by SVD without class labels. Domain labels then measure separation and neighborhood agreement in this representation. Figure~\ref{fig:document-neighborhoods-appendix} displays the documents through independently fitted t-SNE projections~\citep{vandermaaten2008tsne}, while all reported neighborhood statistics are computed in the 50-dimensional space.

Cross-Chunk reaches 70.6\% overall purity, compared with 66.7\% for BatchTopK. Quantitative Biology and Quantitative Finance improve by 9.7 and 3.9 percentage points over the strongest other displayed method. These neighborhoods complement the trained probes in Section~\ref{sec:classification}.

\begin{figure}[t]
\centering
\includegraphics[width=\linewidth]{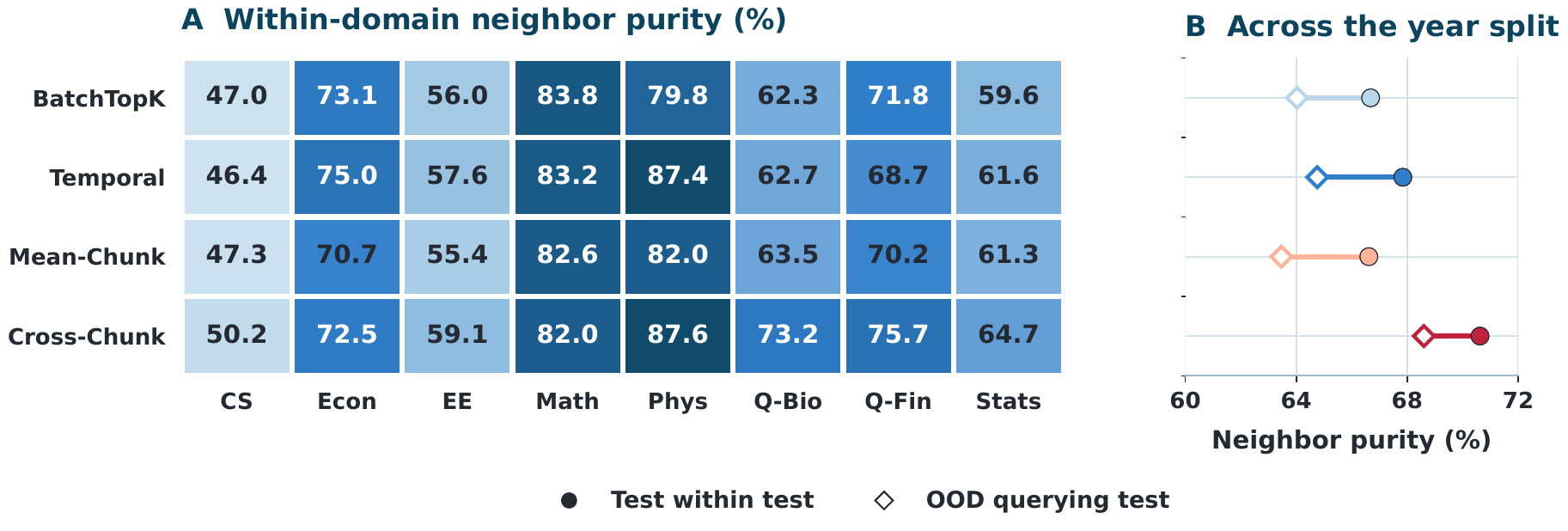}
\caption{\textbf{Semantic neighborhoods across domains and time.} (A) Cosine-neighborhood label purity by ArXiv domain. (B) Filled circles show within-test purity using non-self ranks 2--11; open diamonds show OOD purity over the ten nearest test codes. Both query collections contain 2,048 documents balanced across eight classes. All statistics use label-free 50-D SVD codes.}
\label{fig:semantic-geometry-appendix}
\end{figure}

\begin{table}[ht]
\centering
\small
\caption{Separation, neighborhood agreement, and cross-time nearest-neighbor classification for frozen ArXiv codes. Higher is better; bold marks column maxima across the displayed methods.}
\label{tab:geometry}
\begin{tabular*}{\linewidth}{@{\extracolsep{\fill}}lrrr@{}}
\toprule
Method & Silhouette & Neighbor purity & Cross-time NN \\
\midrule
BatchTopK & 0.086 & 0.667 & 0.745 \\
Temporal & \textbf{0.102} & 0.678 & 0.744 \\
Mean-Chunk & 0.068 & 0.666 & 0.736 \\
Cross-Chunk & 0.100 & \textbf{0.706} & \textbf{0.774} \\
\bottomrule
\end{tabular*}
\end{table}

Silhouette averages $(b_i-a_i)/\max(a_i,b_i)$, where $a_i$ is a document's mean distance to its own class and $b_i$ is its smallest mean distance to another class~\citep{rousseeuw1987silhouettes}. Neighbor purity is the fraction of selected neighbors sharing the query's label. Cross-time NN assigns each later document the majority label among its ten nearest earlier test-set codes, using those codes as its labeled gallery.

For each within-test query, we exclude the query itself and its
closest non-self neighbor, then compute neighborhood purity using
non-self ranks 2--11. For an OOD query, we use the ten nearest codes
from the earlier test set. Both calculations therefore use ten
neighbors per query, but draw them from different reference pools.
Figure~\ref{fig:semantic-geometry-appendix} reports the fraction of
neighbors sharing the query's label, while Table~\ref{tab:geometry}
reports majority-vote accuracy for cross-time classification. These
readouts distinguish local label agreement from the final class
assigned by a neighborhood vote.

Temporal achieves the highest silhouette score, whereas Cross-Chunk
leads in neighbor purity and cross-time nearest-neighbor accuracy.
The rankings can differ because silhouette measures overall class
separation, while the other metrics depend on local neighborhoods.
Economics and Mathematics favor the token baselines; Quantitative
Biology and Quantitative Finance show larger Cross-Chunk gains.
Cross-Chunk's advantage therefore varies across scientific domains.
The domain labels give these neighborhood results a subject-level
interpretation beyond geometric proximity alone.

\section{Classification with Limited Labels}
\label{app:label-efficiency}

\begin{figure}[t]
\centering
\includegraphics[width=\linewidth]{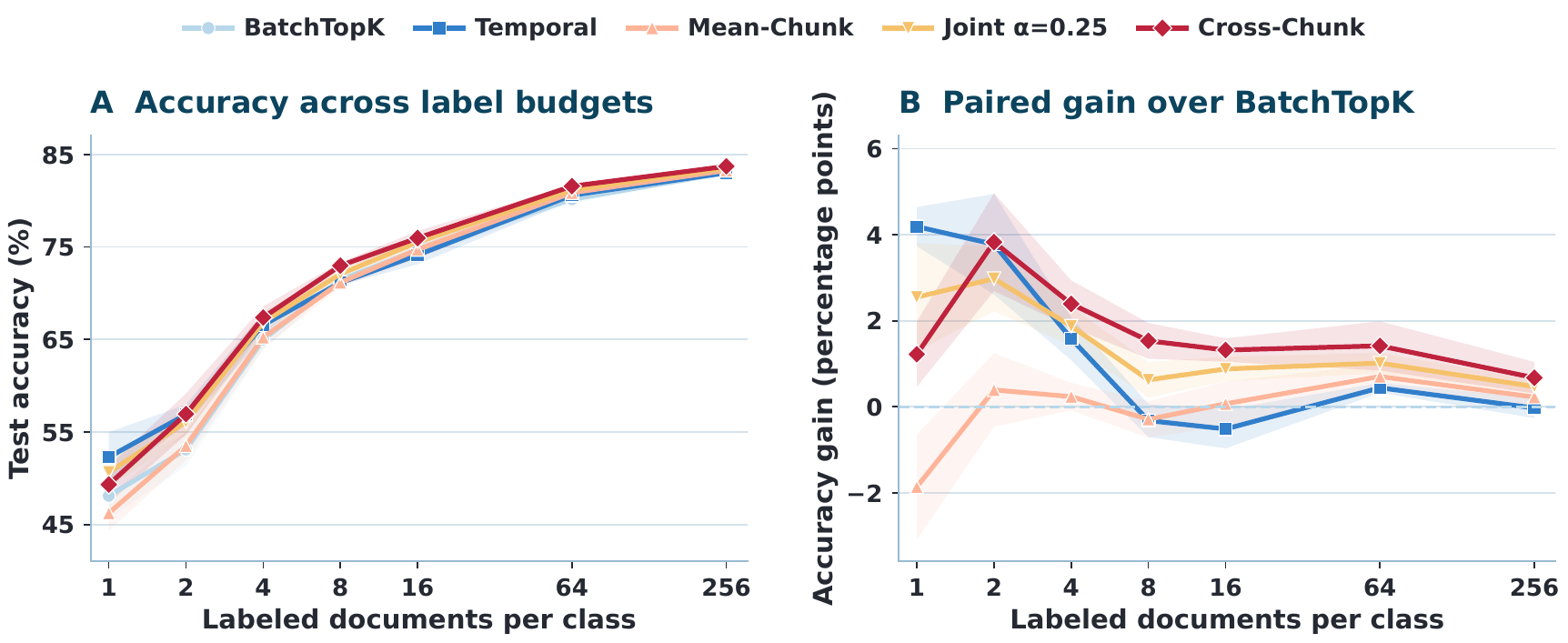}
\caption{\textbf{Classification with limited supervision.} (A) Test accuracy across seven labels-per-class budgets. (B) Within-seed differences from BatchTopK in percentage points. Points average five shared sampling seeds, and bands show one standard error of the mean, computed from paired differences in panel B. The curves expand the aggregate low-label summary in Section~\ref{sec:classification}.}
\label{fig:label-efficiency}
\end{figure}

Frozen-code classification tests whether useful distinctions are accessible to a simple probe as supervision varies~\citep{kantamneni2025useful,karvonen2025saebench}. We use $\{1,2,4,8,16,64,256\}$ labeled documents per class, keeping the representation, probe settings, and evaluation split fixed. Five sampling seeds are used at each budget. All methods receive identical labeled subsets.

Figure~\ref{fig:label-efficiency}A averages test accuracy across these seeds and shows one standard error. Panel B subtracts BatchTopK accuracy within each seed before averaging; its band therefore reflects variation in the paired differences. These curves reveal how performance changes as the same probe receives more examples from each scientific domain. The axis varies labeled examples per class.

Low-label AUC integrates accuracy over the base-two logarithm of the label budget and divides by the observed log-budget range. A doubling occupies one horizontal unit, while the 16-to-64 interval occupies two. Equation~\ref{eq:low-label-auc} provides the trapezoidal calculation. The five seeds vary the labeled subsets supplied to the probes, keeping the trained SAE representations fixed across those classification trials. The curves use one frozen representation per method across all label budgets.

\section{Joint-Chunk Alpha Sweep Across the Evaluation Suite}
\label{app:alpha-suite}

We use $\alpha=0.25$ as a randomly chosen representative Joint-Chunk setting in the main comparisons. The sweep evaluates $\alpha\in\{0.25,0.5,1.0,1.5\}$ with the same architecture, token budget, caches, and downstream procedures. The additional settings test whether the paper's conclusions depend on the particular partner weight used to illustrate Joint-Chunk. All settings share these metric definitions.

The three additional weights retain the central pattern of stronger semantic feature discovery and downstream transfer relative to the token baselines. Their high-level feature fractions range from 29.5\% to 34.4\%, compared with 16.8--17.2\% for the token methods. Reasoning generalization reaches 59.2--64.0\%, compared with 46.6\% for BatchTopK and 25.9\% for Temporal; document retrieval and classification transfer also improve across all three additional evaluated settings.

Changing $\alpha$ adjusts the balance of capabilities. Among the additional settings, $\alpha=0.5$ has the highest high-level fraction and focused interpretability. Larger partner weights reduce reconstruction cosine and native reasoning recall, while retrieval and classification vary less. The main conclusions remain consistent across the tested weights: chunk-level objectives expose semantic structure and support downstream utility, with the observation unit and prediction target determining their complementary strengths. The weight controls the relative emphasis within this shared design.

All Joint-Chunk runs use three random seeds and the shared training protocol. Appendix~\ref{app:ablations} separates self and partner fidelity for each weight, showing how their contributions enter composite RFVE. This target-specific accounting connects the sweep to the main equations while keeping the interpretation of each downstream metric unchanged. The sweep connects weights to feature and task scores.

\section{LLM Judge Prompts and Evaluation Controls}
\label{app:prompts}

Scoring uses DeepSeek V4 Flash at temperature zero~\citep{deepseek2026v4}. DeepSeek V4 Pro generates the reasoning passages and their paired controls. The judge evaluates frozen evidence with SAE names hidden, and the templates use fixed output schemas. The following prompts retain the evaluation instructions; bracketed fields contain the feature-specific descriptions, contexts, or reasoning inputs supplied to each request. Judgments determine the main-text scores.

\subsection{Frozen explanations and high-level scoring}

Explanation generation treats a complete 128-token context as one activation unit. Its instruction asks for a concise semantic or functional pattern rather than a list of isolated trigger words. This choice aligns the explanation with the passage-level interpretation task shared across the five methods. The resulting description is frozen before the judge examines its four active and four inactive scoring contexts. That description is used throughout the fixed scoring panel of complete contexts.

\begin{promptbox}[coltext=black,coltitle=black,title={Explanation generation}]
SYSTEM:\\
We are interpreting one sparse feature over complete 128-token text contexts.
Every example is the same length and is one indivisible activation unit.  The
feature's scalar activation is computed for the whole context.  Summarize one
concise, general pattern shared by the activating contexts: topic, domain,
style, intent, or discourse function.  Do not search for a trigger token, do
not mention activation strength, and do not describe incidental words.  Use at
most 30 words and output only the explanation.\\
USER:\\ Activating contexts: [ACTIVATING CONTEXTS]
\end{promptbox}

The scoring prompt tests whether each complete context matches the frozen description. Its selections determine $r_f^+$ and $r_f^-$ in Eq.~\ref{eq:interp}. All 1,000 sampled features enter the average, including those whose explanations match no active context. The explanation remains frozen across all eight scoring contexts.

\begin{promptbox}[coltext=black,coltitle=black,title={Explanation scoring}]
SYSTEM:\\
We are evaluating a sparse feature over complete 128-token contexts.  Given an
explanation and eight contexts, select the contexts that should activate as a
whole.  Every context is one indivisible unit: do not search for a single
trigger token or substring.  Return only comma-separated 1-based indices, or
None if no context matches.\\
USER:\\ Explanation: [FROZEN EXPLANATION]\\
Contexts: [FOUR ACTIVE AND FOUR INACTIVE CONTEXTS]
\end{promptbox}

The high-level census examines a separate question: whether a randomly sampled coordinate expresses a reusable concept across strong activations. The judge sees ten contexts from distinct held-out documents, with method identity and activation magnitude hidden. The complete rubric records a concept, matched examples, and whether a single surface-only rule is sufficient to explain the observed pattern. The decision rests on content shared across these activating contexts.

\begin{promptbox}[coltext=black,coltitle=black,title={High-level feature census}]
You are classifying one sparse-autoencoder feature from ten typical strong
activations.  The SAE type, feature number, and activation values are hidden.
Identify the most informative common explanation and use exactly one level:
0 means no reusable rule covers at least six examples; 1 means one context-free
surface/form rule alone explains at least eight; 2 means one coherent semantic
or functional concept is the best explanation for at least six.  Surface rules
include exact tokens, fixed phrases, named entities, boilerplate, markup,
punctuation, local syntax, or programming/API spelling.  Related words may
coexist with a level-2 concept; set \texttt{surface\_sufficient=true} only when
a single surface-only rule is sufficient.  Do not use protocol artifacts such
as continuation across a split, generic topical consistency, or one-document
coherence as the feature rule.  Mark exactly the matching examples.  A feature
passes only when level=2, at least six examples match, and
\texttt{surface\_sufficient=false}.  Return one JSON array with id, level,
concept, matching example IDs, \texttt{surface\_sufficient}, and reason.
\end{promptbox}

A passing coordinate receives level 2, matches at least six contexts, and has no sufficient surface-only explanation. The fraction counts these coordinates among all 1,000 sampled features, connecting repeated semantic evidence to the distinction between concepts and surface patterns.

\subsection{Reasoning relations, selection, and calibration}

The reasoning panel contains five relations. Causal mechanism explains how a cause produces an outcome; planning organizes actions toward a goal; backtracking revises a prior choice after an obstacle or new information. Conditional assumption develops consequences under an assumption, and induction extends a pattern across examples. These definitions identify relations expressed by the text, independently of whether a familiar category word or conventional cue phrase appears.

Each relation has 500 native examples: 400 form the discovery pool and 100 are held out. Selection and confirmation require target support and specificity, with a distinct eligible feature assigned to each relation by Eq.~\ref{eq:reasoning-selection}. Feature identities and detection cutoffs are frozen before held-out scoring and before constructing the paired control views, keeping feature discovery separate from the texts used to test generalization. The views share one selected feature per method and relation.

For method $M$, feature $f$, and document $D$, native-resolution scoring takes the maximum over its thresholded token activations or independently encoded chunk activations within the document:
\begin{equation}
S_{M,f}(D)=
\begin{cases}
\displaystyle\max_{t\in D}[\psi_M(\mathbf{h}_t)]_f,
&M\in\mathcal{M}_{\rm token},\\[4pt]
\displaystyle\max_{C\in\mathcal{C}(D)}
[\psi_M(\boldsymbol{\mu}_C)]_f,
&M\in\mathcal{M}_{\rm chunk}.
\end{cases}
\label{eq:reasoning-document-score}
\end{equation}
Here, $\mathcal{C}(D)$ denotes the evaluation chunks in the document. A relation is detected when its selected feature exceeds its fixed cutoff. Calibration uses 208 confirmation-fold Pile documents and sets each cutoff to an empirical background false-positive rate of at most 5\%, giving the subsequent native and control views a common decision rule calibrated against the same background document collection.

\subsection{Separating reasoning relations from surface cues}

A word can accompany a reasoning relation without defining it. For example, ``wait'' can mark a revision of a plan or simply ask someone to remain in place. Testing the relation therefore requires varying its structure and its familiar wording separately. The native, cue-free, and cue-only views implement this distinction: they ask whether a feature recognizes the original relation, recognizes it under new wording, and rejects familiar words when the reasoning relation is absent.

Native passages are generated before the discovery/evaluation split. Each expresses its relation across multiple 128-token chunks with varied domains, entities, and sentence styles. Cue families are used as nuisance controls, and the generator records the relation stages and the cue terms actually present in each passage. The following template fixes the intended structure of these native instances.

\begin{promptbox}[coltext=black,coltitle=black,title={Native reasoning instances}]
Generate exactly [COUNT] independent latent instances for the reasoning
category [CATEGORY], whose defining relation is [DEFINITION].  Use the
registered cue family [CUES] only as a balanced nuisance control: some
instances should contain several cues and some none or a different member;
no cue may be perfectly predictive.  Cover many domains, entities, numeric
ranges, templates, and sentence styles.  Each text must be a natural
paragraph of about 280--420 words (at least 256 tokenizer tokens), with the
relation spread across two or more 128-token chunks.  Do not insert labels
such as Premise, Step, Conclusion, or JSON.  Return only a JSON array.  Each
object must contain exactly these useful fields: \texttt{instance\_id},
\texttt{category}, \texttt{text}, \texttt{relation\_steps} (three concrete
stages), \texttt{cue\_terms} (only cues that actually occur),
\texttt{cue\_present} (boolean), \texttt{domain}, and \texttt{template\_id}.
\end{promptbox}

Two independent requests then create paired control views for each held-out native passage. The cue-free positive preserves the relation, broad topic, and native length stratum while replacing wording and removing cues. The cue-only negative preserves recognizable lexical anchors while replacing the events so that the target relation is absent. Independent rewriting gives each view natural text suited to its intended semantic condition. The views separate relations from their cues.

\begin{promptbox}[coltext=black,coltitle=black,title={Cue-free positive (view B)}]
Write one completely new, independent full-length paragraph for the supplied
source.  Preserve the target reasoning relation, broad topic and domain, and
the native length stratum, but remove every registered cue, observed anchor,
morphological variant, and direct surface equivalent.  Use fresh entities and
sentence structure; do not copy a source sentence, add labels, or mention this
evaluation.  Keep the relation genuinely multi-stage and express it through
ordinary content.  Return only a JSON array with one object containing
\texttt{pair\_id} and \texttt{text}; copy \texttt{pair\_id} exactly.
\end{promptbox}

The negative view retains observed BatchTopK and Temporal anchor words because these supply concrete, activation-linked candidates for a lexical explanation. Retaining such words makes the control informative: a detector must distinguish their use in ordinary content from their use in a genuine reasoning relation. The resulting passages are shared by every method, with frozen features and cutoffs used across all three views. Every method scores identical control views.

\begin{promptbox}[coltext=black,coltitle=black,title={Cue-only negative (view C)}]
Write one completely new, independent full-length paragraph about the same
topic, entities, time frame, and setting, but discard the source events and
target reasoning relation.  Preserve comparable length and include both
literal center-token surfaces supplied for the paired BatchTopK and Temporal
anchors.  Remove the complete relation and all implicit multi-stage
equivalents; do not deny the relation in a meta-sentence, copy the source, or
mention this evaluation.  Return only a JSON array with one object containing
\texttt{pair\_id} and \texttt{text}; copy \texttt{pair\_id} exactly.
\end{promptbox}

Local validation and blind review reject failed rewrites before feature scoring. Cue-free detection supplies positive evidence that the relation remains recognizable after its familiar wording changes. Cue-only rejection tests whether those words are sufficient to produce a detection without the relation. Read together with native recall, these outcomes evaluate reasoning-sensitive feature behavior across the five relation types in both original and independently rewritten passages. The 66\% versus below-5\% backtracking comparison therefore measures recognition under this shared, baseline-anchored lexical stress test, in which cue-free positives remove observed anchors and cue-only negatives preserve token-baseline anchors.

This is because, in real-world retrieval scenarios, it is highly likely that no anchor words are activated. Our experiments must replicate this condition. We believe this comparison is most relevant when SAEs are used to retrieve reasoning examples that lack salient cues such as ``step'' or ``wait''. We do not want the SAE to miss retrieving these inference data, as they are highly valuable. Our chunk-level SAE excels precisely at this, which constitutes the primary significance of this experiment.

\subsection{Matched pooling and cutoff sensitivity}

\begin{figure}[t]
\centering
\includegraphics[width=\linewidth]{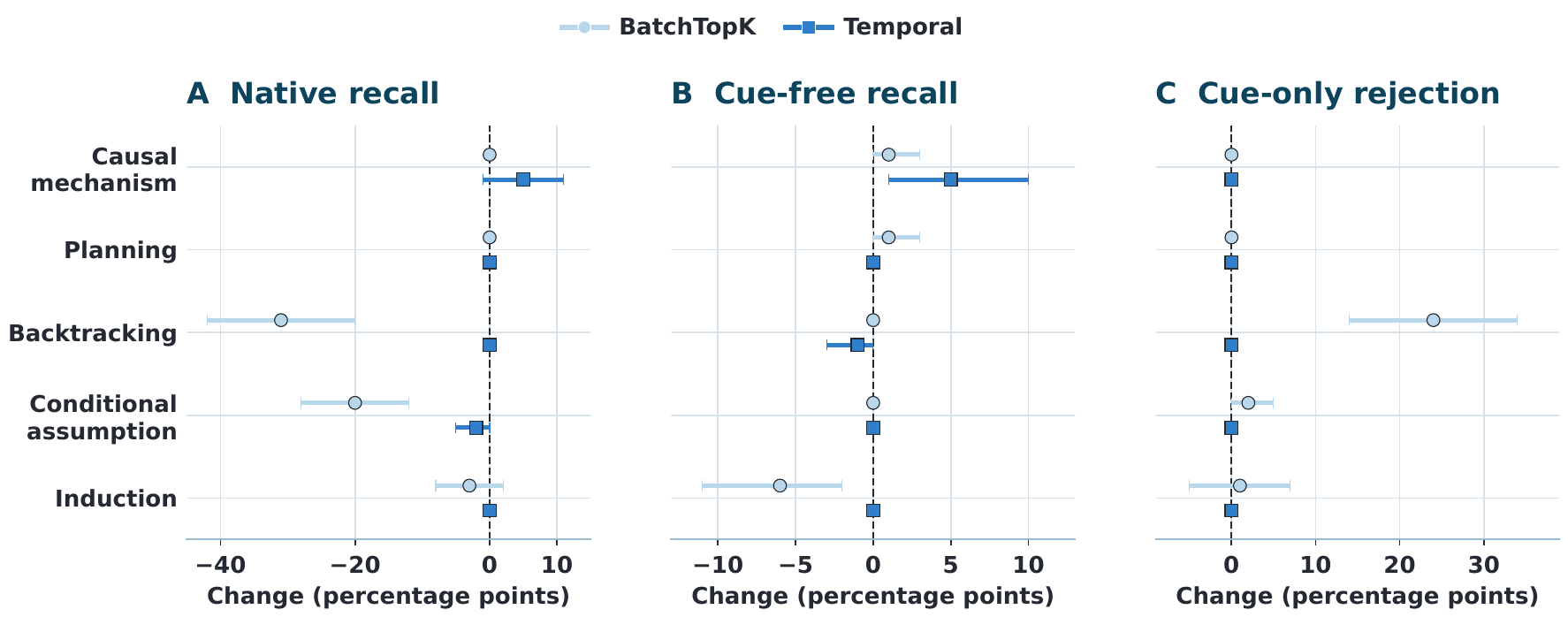}
\caption{\textbf{Reasoning decisions under matched pooling and recalibration.} Matched minus native-resolution rates for BatchTopK and Temporal: native recall, cue-free recall, and cue-only rejection. Comparisons pair the same 100 documents per relation and view; whiskers show paired-bootstrap 95\% intervals from 20,000 resamples. Mean-Chunk and Cross-Chunk retain the same decisions on these views. Positive differences indicate higher reasoning recall or cue-only rejection.}
\label{fig:reasoning-matched-appendix}
\end{figure}

The matched audit pools each token method's thresholded codes within shared 128-token chunks, then takes the maximum chunk score for the document. Feature identities remain fixed. Cutoffs are recalibrated on the same 208 Pile confirmation documents to empirical false-positive rate at most 5\%. This protocol evaluates the joint effect of chunk pooling, document-level score opportunities, and background recalibration on document decisions made by the selected reasoning features.

Figure~\ref{fig:reasoning-matched-appendix} pairs the same 100 held-out documents per relation and view. BatchTopK's backtracking native recall falls by 31 percentage points, while its cue-only rejection rises by 24 points. Temporal changes remain within five points, and Mean-Chunk and Cross-Chunk retain their decisions. The paired-bootstrap intervals describe changes in the document decisions made by these frozen features under the two calibrated scoring protocols. Differences pair documents across protocols.

\subsection{Steering generation and scoring}

Steering tests whether adding a feature's decoder direction moves generation toward its frozen concept description~\citep{wang2025utility,wu2025axbench}. We evaluate 1,000 features per method, including each Joint-Chunk weight. Generation is greedy, with a 192-token prompt and at most 32 new tokens. At every non-padding layer-21 position during prefill and generation, the intervention adds $\gamma\mathbf{d}_f/\|\mathbf{d}_f\|_2$, using the unit-normalized decoder intervention defined in Eq.~\ref{eq:steering}.

Following AxBench's concept-specific steering-strength selection \cite{wu2025axbench}, we evaluate every feature and method on the same prespecified strength grid and report the feature-wise maximum as a measure of attainable steering score. We simplify the process and extract the core ideas. Base strengths are $\{1,3,5,7,10\}$. When the entire continuation remains unchanged at every base strength, the feature enters the ordered rescue grid $\{15,20,30,50\}$. Rescue stops at the first changed continuation or when the grid is exhausted. Baseline and intervention use the same prompt and decoding settings for each feature, preserving a direct comparison between the resulting continuations.

The blinded judge scores concept direction $C_{f,\gamma}$ and coherence preservation $H_{f,\gamma}$ on 0--100 scales. The deterministic combination applies the main-text coefficients to the two judge scores
\begin{equation}
s_{f,\gamma}
=0.7C_{f,\gamma}
+0.3C_{f,\gamma}\frac{H_{f,\gamma}}{100}.
\label{eq:appendix-steering}
\end{equation}
Concept direction measures change relative to baseline, and coherence preservation measures additional degradation caused by intervention. An unchanged continuation has $C=50$ and $H=100$, yielding the neutral score 50. The prompt below specifies the two judgments used in this deterministic combination. The returned scores use the main-text coefficients for concept and coherence.

\begin{promptbox}[coltext=black,coltitle=black,title={Steering judge},fontupper=\normalfont\fontsize{8.9}{10.8}\selectfont]
You are a blind, critical evaluator of causal feature steering.  The concept
hypothesis may be wrong.  The same prompt was decoded twice:
\texttt{BASELINE} has no intervention and \texttt{POSITIVE} adds the normalized
decoder direction at every non-padding layer-21 position during prefill and
generation.  The prompt, hypothesis, and both continuations are supplied below.

\looseness=-1
Score exactly two integer rubrics in [0,100].  For
\texttt{concept\_direction}, 0 means strong movement away from the concept,
25 modest movement away, 50 no meaningful change (including identical output),
75 clear movement toward it, and 100 unmistakable dominant movement toward it.
For \texttt{coherence\_preservation}, compare POSITIVE only with BASELINE:
100 means no additional degradation (or an improvement), 75 mild degradation,
50 clear but interpretable degradation, 25 severe degradation, and 0 unusable.
Do not penalize defects already present in BASELINE or reward lexical copying.
Return JSON only with the two fields and a brief comparative rationale; do not
return an overall score.
\end{promptbox}

For each feature, we retain the highest score among its evaluated strengths, including eligible rescue attempts, with ties resolved in favor of the lower strength. The method score averages these feature-wise maxima over the complete 1,000-feature panel. This aggregation follows Eq.~\ref{eq:steering-score} and uses the same strength-search procedure across the compared dictionaries. The averaging unit is the feature, with one selected score contributed by each coordinate in the evaluated dictionary panel.

\section{Joint-Chunk SAE Alpha Ablation}
\label{app:ablations}

The alpha ablation holds the architecture, shared prefix, activation normalization, and training budget fixed while varying $\alpha\in\{0.25,0.5,1.0,1.5\}$. Each run, one billion training occurrences, dictionary width 65,536, and $K=128$. Table~\ref{tab:alpha} evaluates all selected checkpoints on the same 262,144-occurrence monitor, complementing the evaluation-suite comparison in Appendix~\ref{app:alpha-suite}.

Let $F_{\rm self}$ and $F_{\rm partner}$ be the target-specific SAE FVEs. The self reference has FVE one; the dense partner reference has $F_{\rm ref}^{\rm partner}=0.676758$ on this monitor. The composite score applies weights $1$ and $\alpha$ to the two SAE FVEs and to both target-matched reference FVEs on this monitor
\begin{equation}
\rfve_{\rm joint}(\alpha)
=\frac{F_{\rm self}+\alpha F_{\rm partner}}
{1+\alpha F_{\rm ref}^{\rm partner}}.
\label{eq:appendix-joint-rfve}
\end{equation}
The common task-weight factor $1/(1+\alpha)$ cancels from the ratio. This construction compares the weighted explained variance of the sparse predictions with that of the weighted target-matched references, preserving the same partner emphasis in the numerator and denominator.

\begin{table}[ht]
\centering
\small
\caption{\textbf{Target fidelity across Joint-Chunk weights.} Each setting is evaluated on the shared monitor. Self and partner FVE describe the two prediction targets, and composite RFVE uses Eq.~\ref{eq:appendix-joint-rfve}.}
\label{tab:alpha}
\begin{tabular*}{\linewidth}{@{\extracolsep{\fill}}lrrr@{}}
\toprule
$\alpha$ & Self FVE & Partner FVE & Composite RFVE \\
\midrule
0.25 & 0.9263 & 0.6698 & 0.9355 \\
0.50 & 0.9163 & 0.6701 & 0.9350 \\
1.00 & 0.9045 & 0.6662 & 0.9367 \\
1.50 & 0.8966 & 0.6621 & 0.9378 \\
\bottomrule
\end{tabular*}
\end{table}

\looseness=-1 Self FVE decreases from 0.9263 to 0.8966 across the sweep. Partner FVE remains close between $\alpha=0.25$ and $\alpha=0.5$, then reaches 0.6621 at $\alpha=1.5$. Composite RFVE stays between 0.9350 and 0.9378. Its endpoint increase follows from giving greater weight to partner prediction, whose FVE is closer to its own reference, while the separate columns show the absolute behavior of each target.

\section{Evaluation Metric Definitions}
\label{app:metric-definitions}

\subsection{Fidelity and directional agreement}

For evaluation set $\mathcal{E}$, let $\mathbf{y}_i$ be an unscaled target, $\widehat{\mathbf{y}}_i$ its scaled prediction, and $\overline{\mathbf{y}}$ the fixed training-set target mean. The prediction error and constant-predictor error are evaluated on the same targets
\begin{equation}
\begin{aligned}
E_{\rm SAE}&=\sum_{i\in\mathcal{E}}
\|s\mathbf{y}_i-\widehat{\mathbf{y}}_i\|_2^2,\\
E_{\rm const}&=\sum_{i\in\mathcal{E}}
\|s(\mathbf{y}_i-\overline{\mathbf{y}})\|_2^2.
\end{aligned}
\label{eq:fidelity-errors}
\end{equation}
FVE is $1-E_{\rm SAE}/E_{\rm const}$, measuring improvement over the constant predictor~\citep{gao2024scaling}. RFVE divides this value by reference FVE on the same examples and target. The self reference is the identity map; the partner reference is the dense predictor selected in Appendix~\ref{app:protocol}.

Reconstruction cosine compares prediction and target directions. For nonzero vectors, their normalized inner product is computed using the nonzero scaled prediction and target vectors:
\begin{equation}
\operatorname{Cosine}_i
=\frac{\widehat{\mathbf{y}}_i^\top(s\mathbf{y}_i)}
{\|\widehat{\mathbf{y}}_i\|_2\,\|s\mathbf{y}_i\|_2}.
\label{eq:reconstruction-cosine}
\end{equation}
Self predictions use observed activations as targets, and partner predictions use neighboring chunk means. Positive rescaling leaves cosine unchanged, complementing squared-error fidelity.

\subsection{Feature structure}

For persistence, let $I_f(C)=\mathbf{1}\{[\mathbf{z}(C)]_f>0\}$ indicate that feature $f$ activates in chunk $C$. Let $\mathcal{P}_{\rm same}$ and $\mathcal{P}_{\rm shuf}$ contain adjacent same-document pairs and length-matched pairs shuffled across documents. The two coactivation estimates and their difference after averaging across features are
\begin{equation}
\begin{aligned}
\widehat{p}_{f,q}
&=\frac{1}{|\mathcal{P}_q|}
\sum_{(A,B)\in\mathcal{P}_q}I_f(A)I_f(B),
\qquad q\in\{\mathrm{same},\mathrm{shuf}\},\\
\mathrm{PersistenceLift}
&=\frac{1}{|\mathcal{F}|}\sum_{f\in\mathcal{F}}
\left(\widehat{p}_{f,\rm same}-\widehat{p}_{f,\rm shuf}\right).
\end{aligned}
\label{eq:persistence}
\end{equation}
The set $\mathcal{F}$ contains features meeting the audit's support criterion. Positive lift identifies coactivation associated with same-document continuity beyond the length-matched shuffled comparison.

For utilization, let $M_f$ be the activation mass of feature $f$ over the held-out audit contexts, and let $\mathcal{F}_{\rm alive}$ be the sampled alive-feature set. Normalized masses and entropy-equivalent utilization are
\begin{equation}
p_f=\frac{M_f}{\sum_{g\in\mathcal{F}_{\rm alive}}M_g},
\qquad
\mathrm{Utilization}
=\frac{\exp\!\left(-\sum_{f\in\mathcal{F}_{\rm alive}}p_f\log p_f\right)}
{|\mathcal{F}_{\rm alive}|}.
\label{eq:utilization}
\end{equation}
Natural-log entropy makes the numerator an effective feature count~\citep{hill1973diversity}. Expressed as a percentage, utilization reaches 100\% for equal masses and falls as mass becomes concentrated.

\subsection{Document retrieval and classification}

For query $q$, let $r_q$ be the rank of its correct partner among the length-matched candidates. For the complete query set $\mathcal{Q}$, we average the indicator that the partner's rank is at most $k$
\begin{equation}
\mathrm{Recall@}k
=\frac{1}{|\mathcal{Q}|}\sum_{q\in\mathcal{Q}}
\mathbf{1}\{r_q\leq k\}.
\label{eq:document-recall}
\end{equation}
The figures report $100\,\mathrm{Recall@}5$, averaging equally over queries. Every method ranks the same partners and galleries, using its corresponding representation or lexical similarity score.

OOD accuracy is the percentage of future-year documents whose domain is correctly predicted by the frozen-representation multinomial logistic probe. Low-label AUC summarizes test accuracy across budgets $b_j\in\{1,2,4,8,16,64,256\}$. Writing $x_j=\log_2b_j$ and $a_{s,j}$ for fractional accuracy under sampling seed $s$, we average the normalized log-budget integrals across seeds
\begin{equation}
\mathrm{LowLabelAUC}
=\frac{1}{5}\sum_{s=1}^{5}
\frac{1}{x_7-x_1}\sum_{j=1}^{6}
\frac{a_{s,j}+a_{s,j+1}}{2}(x_{j+1}-x_j),
\qquad x_7-x_1=8.
\label{eq:low-label-auc}
\end{equation}
This normalized trapezoidal area summarizes accuracy over label doublings. Averaging seed-wise areas is equivalent to integrating the mean accuracy curve because both operations are linear.

\subsection{Reasoning selection, recall, and generalization}

For relation $c$, let $p_c^{\rm sel}(f)$ be the weighted selection-fold fraction activating feature $f$, with background and other-relation rates defined analogously. The eligible set $\mathcal{E}_c$ contains features meeting support and specificity requirements in selection and confirmation data and available for assignment to the relation. We select the largest target advantage over background or other-relation activation
\begin{equation}
f_c^\star=\operatorname*{arg\,max}_{f\in\mathcal{E}_c}
\left[p_c^{\rm sel}(f)-
\max\left\{p_{\rm bg}^{\rm sel}(f),
\max_{r\neq c}p_r^{\rm sel}(f)\right\}\right].
\label{eq:reasoning-selection}
\end{equation}
The selected feature and calibrated cutoff $\theta_c$ remain fixed across the held-out views. Its document decision is $D_c(x)=\mathbf{1}\{S_{f_c^\star}(x)>\theta_c\}$, giving the corresponding binary detection.

Let $\mathcal{V}_c^{\rm native}$, $\mathcal{V}_c^{\rm free}$, and $\mathcal{V}_c^{\rm only}$ denote the views for relation $c$. Their detection and rejection rates are
\begin{equation}
\begin{aligned}
R_c^{\rm native}
&=\frac{1}{|\mathcal{V}_c^{\rm native}|}
\sum_{x\in\mathcal{V}_c^{\rm native}}D_c(x),\\
R_c^{\rm free}
&=\frac{1}{|\mathcal{V}_c^{\rm free}|}
\sum_{x\in\mathcal{V}_c^{\rm free}}D_c(x),\\
Q_c^{\rm only}
&=\frac{1}{|\mathcal{V}_c^{\rm only}|}
\sum_{x\in\mathcal{V}_c^{\rm only}}\bigl(1-D_c(x)\bigr).
\end{aligned}
\label{eq:reasoning-rates}
\end{equation}
Native recall averages $R_c^{\rm native}$ over the five relations; generalization averages $(R_c^{\rm free}+Q_c^{\rm only})/2$. Both are percentages, with 100 documents per relation and view giving equal relation weights.

Figure~\ref{fig:reasoning} displays cue-only false activation, $1-Q_c^{\rm only}$. Generalization uses the complementary rejection rate, combining cue-free detection with rejection of cues without the relation.

\subsection{Causal steering and aggregation}

Steering uses the unit-normalized decoder intervention in Eq.~\ref{eq:steering} and the concept/coherence combination in Eq.~\ref{eq:appendix-steering}. Let $\Gamma_f$ contain the positive strengths actually evaluated for feature $f$, including eligible rescue attempts. Averaging the selected feature-wise scores over the sampled panel gives
\begin{equation}
S_{\rm steer}(f)=\max_{\gamma\in\Gamma_f}s_{f,\gamma},
\qquad
S_{\rm steer}=\frac{1}{|\mathcal{F}_{\rm steer}|}
\sum_{f\in\mathcal{F}_{\rm steer}}S_{\rm steer}(f),
\label{eq:steering-score}
\end{equation}
where $|\mathcal{F}_{\rm steer}|=1000$. Each feature contributes one maximum with equal weight, and ties favor lower strength, yielding the dictionary-level steering summary used in the main comparison.

\par
\endgroup
\end{document}